\documentclass[11pt]{amii}

\usepackage{environ}
\usepackage{array}
\usepackage{algorithm}
\usepackage{algpseudocode}
\usepackage{placeins}
\usepackage{etoc}
\usepackage{fontawesome5}

\renewenvironment{algorithm*}[1][]{\begin{algorithm}[p]}{\end{algorithm}}
\AtBeginEnvironment{algorithmic}{\linespread{0.96}\selectfont}


\graphicspath{{figs/}}

\newcommand{\compactfigwidth}{0.7\linewidth}

\newcommand{\quoteattr}[1]{#1}

\newcommand{\teaserbody}{%
  \includegraphics[height=0.217\linewidth,trim={5pt 696pt 8pt 3pt},clip]{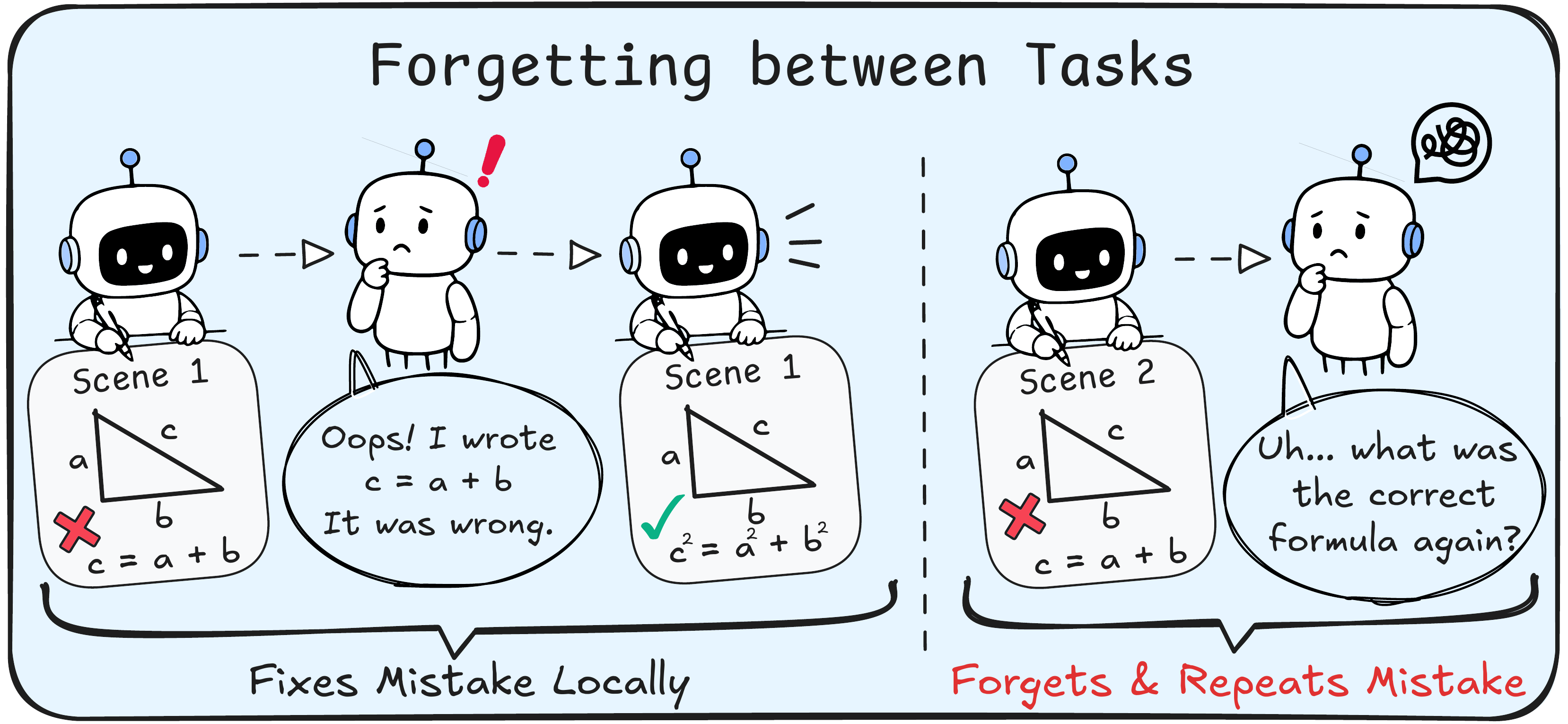}\hfill
  \includegraphics[height=0.217\linewidth,trim={5pt 0pt 8pt 757pt},clip]{motivation.pdf}%
}
\newcommand{\teaserfirst}{Left}
\newcommand{\teasersecond}{Right}

\newcommand{\teaserfloat}{}

\newenvironment{contributions}{\begin{itemize}}{\end{itemize}}

\newcommand{\appendixcontents}{%
  \etocsettocdepth.toc{subsection}%
  \section*{Appendix Contents}%
  \begingroup
  \renewcommand{\contentsname}{}%
  \vspace{-2.2em}%
  \tableofcontents
  \endgroup
  \clearpage}

\RenewEnviron{abstract}{\global\let\abstractlist\BODY}

\urldef{\projectlink}\url{https://manimagent.github.io/}
\urldef{\codelink}\url{https://github.com/jwj1342/Paper2Manim}

\title{ManimAgent: Self-Evolving Multimodal Agents for Visual Education}
\author[1,6,*]{Wenjia~Jiang}
\author[2,*]{Zongyuan~Cai}
\author[2]{Yuanhang~Shao}
\author[3]{Chenru~Wang}
\author[2]{Boyan~Han}
\author[4]{Zhixue~Song}
\author[5]{Keyu~Chen}
\author[3]{Shengwei~An}
\author[2]{Xu~Yang}
\author[1,6,\ddag,\dag]{Zhou~Yang}
\affiliation[1]{University of Alberta}
\affiliation[2]{Southeast University}
\affiliation[3]{Virginia Tech}
\affiliation[4]{Emotion Machine Inc.}
\affiliation[5]{Vivavia Inc.}
\affiliation[6]{Alberta Machine Intelligence Institute (Amii)}
\affiliation[\ddag]{Canada CIFAR AI Chair}
\affiliation[\dag]{Corresponding author}
\affiliation[*]{Equal contribution}
\renewcommand\metadataformat[2][]{{\small \ifstrempty{#1}{}{{\bfseries #1:} }#2}}
\metadata{\faEnvelope~\email{zy25@ualberta.ca}, \email{wj8@ualberta.ca}}
\metadata{\faGlobe~\projectlink}
\metadata{\faGithub~\codelink}
\metadata[Venue]{EMNLP 2026 Main Conference}
\date{August 20, 2026}
\renewcommand{\runningheader}{EMNLP 2026}

\begin{document}
\begin{abstract}
Multi-round reflection lets agents built on large language models recover from failures
within a single task, but lessons learned across many reflection rounds on one
task are discarded before the next begins. We study this gap on a
code-generation task: from a scientific paper section, the agent writes executable
\textsc{Manim} code. We present \textsc{ManimAgent}, a self-evolving multimodal
agent that carries reflection experience across tasks through a dual-channel
Episodic Memory Bank grown entirely from its own task stream, with no weight
updates and no human seeds. After each animation converges, a vision--language
model scores the rendered keyframes; the resulting signals populate a positive
channel $\mathcal{M}^{+}$ that stores success rationales as soft Reference
Examples, and a negative channel $\mathcal{M}^{-}$ that stores VLM-screened
failure patterns as hard Known Pitfalls. On a fixed-probe evaluation against
no-memory, matched-budget retrieval-augmented generation, and shuffled-memory
baselines, blind human Pass@1 rises and reflection rounds fall as memory size
grows.

\end{abstract}
\maketitle
\vspace{-10pt}
\begin{figure}[ht]\input{sections/01_teaser}\end{figure}

\etocsettocdepth.toc{none}
\section{Introduction}
\label{sec:intro}

\begin{quote}
\small\itshape
\begin{center}
``Memory is not an instrument for exploring the past but for exploring the future.''\\*
\upshape --- \quoteattr{Endel Tulving}
\end{center}
\end{quote}

\teaserfloat

Large language models and their agentic wrappers have become capable enough to handle complex,
multi-stage code generation tasks~\citep{chen2021evaluating,yao2023react,NEURIPS2023_1b44b878}. A recent line of work
applies this capability to visual education through
\textsc{Manim}~\citep{The_Manim_Community_Developers_Manim_Mathematical_2026},
a Python animation library that programmatically renders mathematical and pedagogical scenes from
declarative scripts. \textsc{Manim} was originally developed for the 3Blue1Brown YouTube channel
and is now community-maintained. Concretely, a multi-agent system reads a section of a scientific
paper and writes \textsc{Manim} code that renders a short teaching animation, with promising
results on isolated tasks~\citep{chen2026code2video,jain2025manimator,joshi2026llm2manim}.

However, these systems are forgetful. A state-of-the-art reflection pipeline asked to animate the
Fourier transform may spend many reflection rounds, and thousands of tokens, learning that a
formula must be left-aligned, that a $0.3$~s transition is too fast, or that a specific
\LaTeX{} macro silently breaks under \textsc{Manim}; the animation is delivered, and every one of
those lessons is thrown away. When the same system is later asked to explain a probability
distribution, it makes the same layout mistakes and pays the full reflection cost again.
Reflection-based agents~\citep{NEURIPS2023_91edff07,NEURIPS2023_1b44b878} improve repeated
trials through feedback, but do not evaluate whether visual-programming experience transfers
across disjoint tasks. We call this structural failure cross-task forgetting.

Intuitively, a competent human educator would not start from zero on every new topic. An
instructor who has explained Fourier transforms many times carries two kinds of accumulated
experience: positive templates of how the time-frequency picture should be staged, and negative
lessons about the formula and the diagram overlapping until the audience loses the thread. The
first kind tells the educator what to do; the second kind tells them what to
avoid. Cross-task forgetting, in this light, is the absence of either kind of memory across
episodes. Motivated by this analogy, we extend the standard reflection pipeline with an
explicit cross-task memory of the same two flavours. Concretely, we present \textsc{ManimAgent},
a self-evolving multi-agent system that narrows the inter-task gap by adding two components to
a standard reflection pipeline: a vision--language model (VLM) acting as a structured reward source over
rendered keyframes, and a dual-channel Episodic Memory Bank (EMB) grown entirely from the
task stream of the system itself, with no human seeds and no weight updates.

The first component is a strong VLM, drawing on LLM/VLM-as-a-judge evaluation
work~\citep{liu2023g,NEURIPS2023_91f18a12,lee2024prometheus}, used as a structured reward source. Given the
rendered keyframes, the source teaching text, and the storyboard, the VLM returns a multi-axis
critique covering logical flow, layout and occlusion, and pedagogical accuracy. This critique
plays two roles inside \textsc{ManimAgent}: it drives the per-task reflection loop in which the
system iteratively revises the generated code, and it gates the memory-write step that
decides what is added to the EMB at the end of each task.

The second component is the Episodic Memory Bank itself, a dual-channel external store that the
system populates as it processes a stream of paper-section tasks. After each task converges, two
complementary signals are distilled. The positive channel $\mathcal{M}^{+}$ records a
natural-language success rationale attached to every high-scoring scene, acting as a soft global
template for what good looks like on related tasks. The negative channel $\mathcal{M}^{-}$
  records a structured VLM-screened failure pattern for every improving
  \textsc{failure}$\rightarrow$\textsc{success} transition in the reflection trace, acting as a
  hard exclusion list that rules out fatal mistakes the positive channel alone cannot surface
  (the operational VLM-score write gate is detailed in
\S\ref{sec:method-emb}). On subsequent tasks, both channels are retrieved by task embedding and
injected into the Coder prompt as Reference Examples and Known Pitfalls.

To validate this design, we grow the EMB on a memory-building stream and then
freeze snapshots of different sizes. Each snapshot is evaluated on the same
disjoint probe tasks in read-only mode, so probe outputs cannot be written back
to memory and task order cannot explain differences between conditions
(\S\ref{sec:rq1}). Because the same VLM both scores rendered keyframes and gates memory
writes, using VLM scores as the headline metric would risk evaluator leakage (the
reward model is also the judge, so the system could appear to improve simply by drifting
toward the preferences of the VLM). We therefore use blind human judgement of first-attempt
videos together with reflection rounds as the headline evidence, and report VLM scores only
as auxiliary diagnostics (\S\ref{sec:rq3}).

This work makes three contributions.%
\begin{contributions}%
\item We propose \textsc{ManimAgent}, a self-evolving
multi-agent system that narrows the inter-task gap with a VLM reward source and a
dual-channel EMB, with no weight updates (\S\ref{sec:method}).
\item Two LLM distiller
agents convert each reflection trace into a free-form success rationale
  ($\mathcal{M}^{+}$) and a structured failure lesson ($\mathcal{M}^{-}$), gated by
  VLM-scored improvement (\S\ref{sec:method-emb}).
\item We will also release a JSON-indexed
paper-section animation dataset with scene-role labels and an output-level
human-scoring protocol, and evaluate on it via a fixed-probe snapshot design that
isolates cross-task memory from task-order effects~
(\S\ref{sec:rq1},~\S\ref{sec:exp-data}).%
\end{contributions}

\section{Related Work}
\label{sec:related}
\label{sec:related-work}

\paragraph{Retrieval-augmented generation and episodic memory in LLM agents.}
Retrieval-augmented generation retrieves external evidence into the prompt of a model
instead of storing all task knowledge in parameters~\citep{NEURIPS2020_6b493230}. For code
generation, systems such as DocPrompting and RepoCoder retrieve documentation, API
references, or repository context to improve generation
quality~\citep{zhou2023docprompting,zhang2023repocoder}. Episodic-memory agents such as
MemGPT, Generative Agents, and ExpeL write experiences back to persistent stores during or
after execution~\citep{packer2023memgpt,park2023generative,zhao2024expel}. We share
the retrieval interface with these lines and adopt a matched-budget Manim-code RAG corpus
(\S\ref{sec:exp-setup}) as a strong baseline. However, our memory is self-generated from
VLM-scored animation traces rather than from static documents or unconstrained
  conversation logs; it is written only under explicit operational write gates; and it is
split into positive and negative channels so that reference examples and known pitfalls can
play different roles in the Coder prompt. Appendix~\ref{sec:appendix-emb-vs-rag} makes this
contrast explicit.

\paragraph{In-context example configuration.}
Multimodal in-context learning studies show that demonstration selection,
quality, and composition can substantially affect model behaviour. Yang et
al.~\citep{NEURIPS2023_804b5e30} study image selection and caption configuration
for image captioning, while Li et al.~\citep{Li_2024_CVPR} extend demonstration
retrieval and configuration to visual question answering. \textsc{ManimAgent}
is adjacent to this literature because retrieved experiences likewise provide
prompt-time guidance, but it is not formulated as a multimodal ICL method: its
examples are distilled from the agent's own historical trajectories, admitted
through VLM/reward-gated memory writes, and separated into positive Reference
Examples and negative Known Pitfalls.

\paragraph{Lifelong learning and skill libraries.}
Voyager is the closest prior work in spirit: an LLM-driven agent in \textsc{Minecraft}
accumulates an external skill library that subsequent tasks can call~\citep{wang2023voyager}.
Related web-agent work also uses large multimodal models to act across long-horizon
interfaces~\citep{he2024webvoyager}. \textsc{ManimAgent} differs in three ways. First, the
reward signal is not a sparse rule-based success bit, but a continuous, multi-axis visual
  judgement from a VLM. Second, the memory is dual-channel: positive rationales and VLM-screened
  failure patterns are distilled with explicit VLM-score improvement gating
(\S\ref{sec:method-emb}), whereas skill-library agents mainly preserve successful solutions.
Third, the EMB is grown entirely from the task stream of the system itself, with no seed primitives,
human-curated examples, or external video corpus. Recent Manim-generation work also explores
agentic inference strategies for animation generation~\citep{silva2026training}; our
focus is complementary, on cross-task memory and fixed-probe evaluation.
Appendix~\ref{sec:appendix-comparison} (Table~\ref{tab:comparison}) summarises these
distinctions, and Appendix~\ref{sec:appendix-related-extended} extends the related-work
discussion to programmatic animation pipelines and reflection-based agents.

\section{Method}
\label{sec:method}

\begin{figure*}[t]
  \centering
  \includegraphics[width=\linewidth]{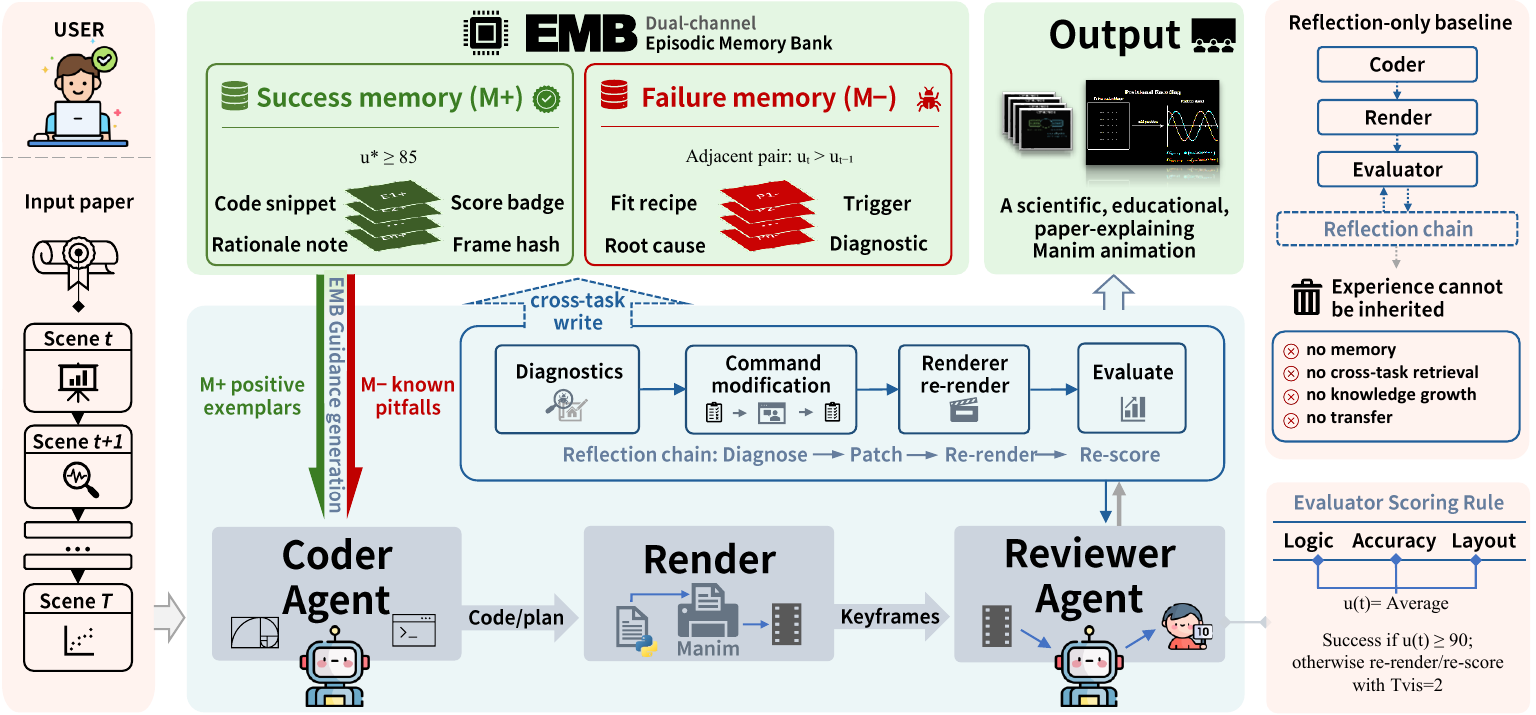}
 \caption{\textbf{\textsc{ManimAgent} closes a VLM-guided repair loop with
 dual-channel cross-task memory.} The visual reviewer uses score $u(t)$ for
 auto-pass and revision control, while the selected score $u^{\ast}$ and
   VLM-scored improvement transitions govern separate success- and failure-memory
 writes.}

  \label{fig:pipeline}
\end{figure*}

\textsc{ManimAgent} produces a short \textsc{Manim} teaching animation from a
paper section (text, scene role, domain tag) via a standard reflection
pipeline of six agents that storyboard, code, render, diagnose crashes, score
keyframes with a VLM, and revise. To this pipeline we add a self-grown,
VLM-gated, dual-channel Episodic Memory Bank (EMB) and a pair of LLM
distillers: the positive channel $\mathcal{M}^{+}$ stores success
rationales injected as soft Reference Examples, and the negative
  channel $\mathcal{M}^{-}$ stores VLM-screened failure patterns injected as hard
Known Pitfalls. Both channels are distilled at consolidation time,
so the EMB holds transferable lessons rather than literal before/after diffs.
Across the task stream, the system reads from the EMB before generating and
writes to it after converging, with no weight updates and no human seeds
(Figure~\ref{fig:pipeline}).

\subsection{Setup and Notation}
\label{sec:method-overview}

A task is a paper section $\tau=(s,r,d)$, where $s$ is the section text, $d$ is its domain
tag, and the scene role $r$ takes one of \textsc{background}, \textsc{method},
\textsc{experiment}, or \textsc{conclusion}. The \textsc{Storyboarder} $\Sigma$ decomposes
$\tau$ into ordered scene specifications $\{\sigma_i\}_{i=1}^{n}$; each $\sigma_i$ carries a
scene name, the claim to be conveyed, the evidence to use, and the final takeaway, which
downstream agents are required to preserve. Within a scene, $t$ indexes reflection iterations inside whichever loop is active
(text-reflection or visual-reflection, defined in \S\ref{sec:method-loop});
$c_i^{(t)}$ is the candidate \textsc{Manim} code at iteration $t$, $K_i^{(t)}$ are $n_K{=}4$ keyframes
sampled from a successful render, and $u(t)\in[0,100]$ is the aggregate VLM score on
those keyframes (defined only after a render succeeds, since text-reflection iterations may
crash before rendering). The execution trace of a scene records the tuple
$(c_i^{(t)}, \text{render result}, u(t)\text{ if available}, \text{diagnostic})$ at every
iteration; consolidation reads from this trace. A frozen sentence encoder $\phi(\cdot)$
embeds retrieval~queries.

The external memory is a single store $\mathcal{M}=\mathcal{M}^{+}\cup\mathcal{M}^{-}$ backed
by one SQLite table and one retrieval interface, but indexed by two per-polarity
Faiss \texttt{IndexFlatIP} structures so that a query against $\mathcal{M}^{+}$ never returns
a record drawn from $\mathcal{M}^{-}$ even at high recall. Records share a context header
$(s_\sigma, r_\sigma, d_\sigma, \mathrm{paper\_id}, \mathrm{section\_id})$, used to form
the retrieval query, and a provenance block (run id, scene id, extraction source, transition
ordinal, before/after scores) used for write-time deduplication; they differ in body fields
by polarity (\S\ref{sec:method-emb}, Table~\ref{tab:emb-schema}).

\subsection{The Self-Evolving Loop}
\label{sec:method-loop}

For each scene $\sigma_i$, the loop has three nested layers: a memory retrieve call, a
text-reflection loop that handles render crashes, and a visual-reflection loop
that handles renderable-but-pedagogically-weak output. The two reflection loops are driven
by different agents, have independent budgets, and write to the EMB through different paths.
We describe each layer in turn; the full pseudocode is in
Appendix~\ref{sec:appendix-pseudocode} (Algorithm~\ref{alg:self-evolving-loop}).

\paragraph{Retrieve.}
For scene $\sigma_i$ in task $\tau$, we form the retrieval query $e_i=\phi([s;r])$ using
all-MiniLM-L6-v2~\citep{reimers2019sentence,sentence_transformers_allminilm_l6_v2}
and fetch the top-$k_{+}=2$ entries $E^{+}$ from $\mathcal{M}^{+}$ and
top-$k_{-}=3$ entries $E^{-}$ from $\mathcal{M}^{-}$ via Faiss
\texttt{IndexFlatIP}~\citep{johnson2019billion,douze2026faiss}
($L_{2}$-normalised vectors, so inner product equals cosine). The two result sets fill
disjoint prompt slots: Reference Examples for $E^{+}$ as soft guidance, and
Known Pitfalls for $E^{-}$ as hard constraints. When the bank is empty, both slots
carry an explicit [No entries available] marker, so prompt structure stays invariant
across EMB sizes and any quality difference is attributable to retrieved content rather than
to formatting changes.

\paragraph{Generate and repair.}
The \textsc{Coder} conditions on $\sigma_i$, $E^{+}$, and $E^{-}$ and emits an
initial program $c_i^{(0)}$ in a single LLM call; subsequent iterations
regenerate the full \textsc{Manim} \texttt{Scene} subclass from scratch with
new feedback appended, because patching a broken script tends to inherit the
same buggy structure. A static checker rejects unsafe imports and calls before
render, and the \textsc{Renderer} executes each candidate inside a sandbox.
On a crash, a text-only \textsc{Reviewer} emits a short hint and a
\texttt{retry}/\texttt{give\_up} decision; the text-reflection loop is bounded
by $T_{\mathrm{text}}{=}2$ retries and short-circuits when the same error
category recurs.

\paragraph{Visual reflection and best-of-$N$ delivery.}
Once a render succeeds, control passes to a separate visual-reflection loop
driven by a \textsc{VLM Reviewer} that scores keyframes on three pedagogical
axes (logical flow, layout/occlusion, accuracy); a distinct
\textsc{Visual Reviser} agent then rewrites the script while preserving the
claim, evidence, and final takeaway of the scene. The loop terminates by VLM
verdict, by an auto-pass override at $u(t)\ge\theta_{\mathrm{conv}}=90$, or at
$T_{\mathrm{vis}}{=}2$. Across the $\le T_{\mathrm{vis}}{+}1$ renderable
candidates we ship the rendition $c^{\star}$ with the highest aggregate score
$u^{\ast}=\max_{t}u(t)$ rather than the last candidate, guarding against
polishing-pass regressions. Separating crash repair from visual revision is
deliberate: the two failure modes have different prompts and budgets, and
conflating them risks one budget swallowing the other. Sandbox limits, short-circuit and tie-break rules, and
distillation length caps are listed in
Appendix~\ref{sec:appendix-impl-details}.

\subsection{Dual-Channel Memory Design}
\label{sec:method-emb}

Two choices define the EMB: maintaining two channels rather than one, and gating each
channel with a different write rule. We motivate both; the full field-level schema is
deferred to Appendix~\ref{sec:appendix-emb-examples} (Table~\ref{tab:emb-schema}).

\paragraph{Why two channels, not one.}
$\mathcal{M}^{+}$ records what a good solution looks like; $\mathcal{M}^{-}$ records what to
avoid. A positive-only skill-library architecture (as in Voyager-style agents~\citep{wang2023voyager}) has no
mechanism for hard exclusion: a known-bad pattern would have to be encoded indirectly, by
hoping that positive exemplars outvote it at retrieval time. Keeping the channels separate
lets us inject $\mathcal{M}^{+}$ as a soft prior and $\mathcal{M}^{-}$ as a hard rule,
mirroring the shape of cue a human educator carries: positive templates plus explicit
pitfalls.

\paragraph{Positive-channel write rule.}
A scene whose best-of-$N$ rendition $c^{\star}$ scores $u^{\ast}\ge\theta_{\mathrm{high}}=85$
is written to $\mathcal{M}^{+}$ as a success rationale plus the chosen code, score, and frame
hash. We deliberately set $\theta_{\mathrm{high}}<\theta_{\mathrm{conv}}$,
so that every scene which triggers the auto-pass also clears the write bar, and a scene that
narrowly converged in the last visual-loop iteration is still remembered. The threshold makes
$\mathcal{M}^{+}$ a high-precision soft prior: every retrieved exemplar has cleared an
explicit quality bar.

\paragraph{Two kinds of failure transition.}
The execution trace defined in \S\ref{sec:method-overview} contains two structurally
different improvement signals, which we distil
separately. Text transitions are crash$\to$success pairs at adjacent text-reflection
iterations: $c_i^{(t-1)}$ raises an error of category $\kappa$, $c_i^{(t)}$ renders cleanly.
Since there is no continuous score across a crash, we accept every such transition and use
the error category plus traceback tail as the diagnostic. Visual transitions are $v_{\mathrm{rev}}\to v_{\mathrm{rev}}+1$ pairs at
adjacent visual-reflection iterations: both render cleanly, both receive aggregate VLM
scores, and we require $u(t)-u(t-1)\ge\Delta_{\mathrm{fail}}=5$ on the $0$--$100$
scale, using the \texttt{revision\_instruction} as the diagnostic. The text-margin
requirement is degenerate (a crash has no numeric score); the visual margin filters out
  noise from a stochastic reviewer.

\paragraph{Lessons are LLM-distilled, not literal diffs.}
For each surviving failure transition, a separate \textsc{Lesson Distiller}
agent consumes the before/after code and the diagnostic and emits the six
$\mathcal{M}^{-}$ body fields; for success records, a \textsc{Rationale Writer}
agent consumes the chosen code and emits a high-level rationale $\rho$. We use LLM
distillation rather than literal diffs because raw before/after code is
dominated by scene-specific tokens (variable names, numerical constants,
camera angles) that would mislead retrieval; the distillation step lifts the
lesson $\ell$ to a transferable description. Field-level length caps are listed in Appendix~\ref{sec:appendix-impl-details}.

\paragraph{Provenance, deduplication, and embedder pinning.}
The dedup key on the SQLite store combines run id, scene id, polarity, extraction source, and
transition ordinal. The ordinal lets $v_{0}\to v_{1}$ and $v_{1}\to v_{2}$ remain distinct
records rather than the later overwriting the earlier, so a long reflection
trace contributes multiple lessons when it should. The embedder identity (model, dim,
version) is pinned in an on-disk spec file at first creation and refused on reopen if a
different encoder is requested, so EMB snapshots cannot be silently re-embedded between
writes and reads, a precondition for the fixed-probe protocol of \S\ref{sec:rq1}, which
evaluates the same snapshot multiple times. At evaluation time, an \texttt{--emb-readonly}
flag nulls out the post-task consolidation step so that the reflection traces of the probe set
cannot leak into the snapshot under test.

\paragraph{No capacity, no eviction.}
We do not currently apply capacity bounds, eviction policies, or paraphrase-level
deduplication: $\mathcal{M}^{+}$ and $\mathcal{M}^{-}$ grow monotonically across the
memory-building stream. \texttt{hit\_count} and \texttt{last\_used} columns are recorded in
provenance for future consolidation work but are not consumed by the current retrieval
interface. All numerical hyperparameters introduced above are summarised in
Appendix~\ref{sec:appendix-reproducibility} (Table~\ref{tab:fixed-hparams}); the conceptual
contrast between EMB and standard retrieval-augmented generation is deferred to
Appendix~\ref{sec:appendix-emb-vs-rag}.

\section{Experiments}
\label{sec:experiments}

\begin{table*}[!t]
\centering
\footnotesize
\setlength{\tabcolsep}{4pt}
\renewcommand{\arraystretch}{1.05}
\newcommand{\groupcap}[1]{\textcolor{blue!55!black}{\textbf{#1}}}
\begin{tabular}{@{} l p{0.44\textwidth} c c c @{}}
\toprule
\textbf{Condition / Variant} & \textbf{Intervention or note}
  & \shortstack{Human\\Pass@1 $\uparrow$}
  & \shortstack{Reflection\\rounds $\downarrow$}
  & \shortstack{Human\\quality $\uparrow$} \\
\midrule
\multicolumn{5}{@{}l}{\groupcap{Headline conditions}\,---\,fixed-probe evaluation under blind human scoring} \\
\cmidrule(lr){1-5}
VLM reflection only              & Standard reflection without retrieval                       & 68.5                  & 12.2                 & 3.41                  \\
Manim-code RAG                   & Static curated Manim corpus, matched retrieval              & 83.3                  & 11.4                 & 3.65                  \\
Random EMB                       & Shuffled EMB content, matched count and budget              & 69.6                  & 11.3                 & 3.42                  \\
EMB@0                            & Empty self-grown memory                                     & 62.0                  & 12.2                 & 3.26                  \\
EMB@50                           & 50-entry self-grown EMB snapshot                            & 66.7                  & 11.3                 & 3.31                  \\
EMB@100                          & 100-entry self-grown EMB snapshot                           & 75.0                  & 10.8                 & 3.43                  \\
EMB@200 (= Full EMB)             & 200-entry EMB snapshot, both channels gated                 & $\mathbf{84.9}$ & $\mathbf{6.5}$ & $\mathbf{3.88}$ \\
\addlinespace[4pt]
\multicolumn{5}{@{}l}{\groupcap{(a) Memory channels}\,---\,which polarity carries the signal?} \\
\cmidrule(lr){1-5}
No $\mathcal{M}^{+}$ (success)   & Retrieve only failure memories                              & 70.0                  & 12.0                 & 3.40                  \\
No $\mathcal{M}^{-}$ (failure)   & Retrieve only success memories                              & 80.9                  & 11.6                 & 3.52                  \\
No memory                        & Disable both memory channels                                & 61.8                  & 12.2                 & 3.20                  \\
\addlinespace[4pt]
\multicolumn{5}{@{}l}{\groupcap{(b) Memory content and retrieval}\,---\,what gets stored and looked up?} \\
\cmidrule(lr){1-5}
Rationale-only $\mathcal{M}^{+}$ & Drop code excerpt from positive injection                   & 75.0                  & 10.5                 & 3.50                  \\
Ungated $\mathcal{M}^{-}$        & Store every diagnostic, no improvement gate                 & 81.8                  & 8.4                  & 3.64                  \\
Top-$1$ retrieval                & One neighbour per channel ($k_{+}{=}k_{-}{=}1$)             & 81.0                  & 9.9                  & 3.84                  \\
\addlinespace[4pt]
\multicolumn{5}{@{}l}{\groupcap{(c) Reviewer and static-corpus controls}\,---\,gains beyond reward and retrieval per se} \\
\cmidrule(lr){1-5}
Weak VLM                         & Step-3.7-Flash replaces the GPT-5.5 reviewer                & 80.0                  & 8.6                  & 3.42                  \\
RAG (doc only)                   & Manim docs only, matched budget                             & 82.1                  & 11.5                 & 3.57                  \\
\bottomrule
\end{tabular}
\caption{\textbf{Headline conditions and ablations on the fixed-probe set share a single human-scoring protocol.} The top block reports the seven
headline conditions used in \S\ref{sec:rq1}; EMB@200 (= Full EMB) is best on
all three metrics and the only condition that improves over Manim-code RAG
on all three headline metrics. The lower three blocks ablate Full EMB: (a) isolates the
contribution of each polarity, (b) stresses the consolidation pipeline
(rationale vs.\ code, gate stringency, retrieval depth), and (c) separates EMB
gains from reviewer strength and from a static documentation-only retrieval
corpus. RAG (doc + code) is identical to the headline ``Manim-code RAG'' row
and not repeated.}
\label{tab:main-results}
\label{tab:ablations}
\end{table*}

To validate the effectiveness of \textsc{ManimAgent}, we conduct experiments on our
paper-section animation dataset with frozen EMB snapshots, matched-budget RAG and
Random-EMB baselines, and a blind human-scoring protocol. The central question is whether a
self-grown Episodic Memory Bank (EMB) makes a VLM-reflection animation agent improve across
tasks without any model-weight updates. Because the same VLM diagnoses rendered keyframes and
gates memory writes, we do not use VLM scores as the primary evidence. Instead, the headline
metrics are blind human judgements on first-attempt outputs and the reflection rounds of the system;
ablations on channel isolation, write gates, retrieval depth, and VLM--human agreement address
mechanism and reward reliability.

\subsection{Evaluation Task Stream}
\label{sec:exp-data}

Our evaluation instrument is a released, indexed paper-section dataset. Each
task asks the system to read a local unit of a scientific paper and generate a
short \textsc{Manim} animation for that unit. The release separates tasks at the
paper level into three roles: \textbf{memory-building} (39 tasks) grows the EMB
before snapshots are frozen; \textbf{fixed-probe} (33 tasks) evaluates each
frozen snapshot in read-only mode and supplies the headline metrics;
\textbf{cross-test} (40 tasks) is a held-out exploratory split used only for the
qualitative inspection of retrieved memories in
Appendix~\ref{sec:appendix-qualitative}. Model-visible inputs are restricted to
the paper title, abstract, target-unit excerpt, required prior context, prior
objects, scene role, domain, and source type; evaluation-only fields (key
claims, reference scene plans, rubrics, condition identifiers, fatal flags) and
output-level human scores are kept outside the model-input subtree and never
written to the EMB. A quarantined debugging holdout, the field-isolation
discipline, and the release validator are detailed in
Appendix~\ref{sec:appendix-validator} and Appendix~\ref{sec:appendix-dataset-summary}.

Starting from an empty memory bank, we process the memory-building stream and
freeze snapshots at
\[
\mathcal{K}=\{0,50,100,200\}.
\]
During probe evaluation, the snapshot is used only for retrieval; no probe
output is written back to memory. Thus, differences between EMB@0 and EMB@$K$
on the same probe tasks measure accumulated cross-task memory rather than
test-time learning. The task scope is local paper-section animation; full-paper
video generation is left outside the headline experiment to avoid narrative and
long-video confounds.

\subsection{Experimental Setup}
\label{sec:exp-setup}

\begin{figure*}[!t]
\centering
\includegraphics[width=\linewidth]{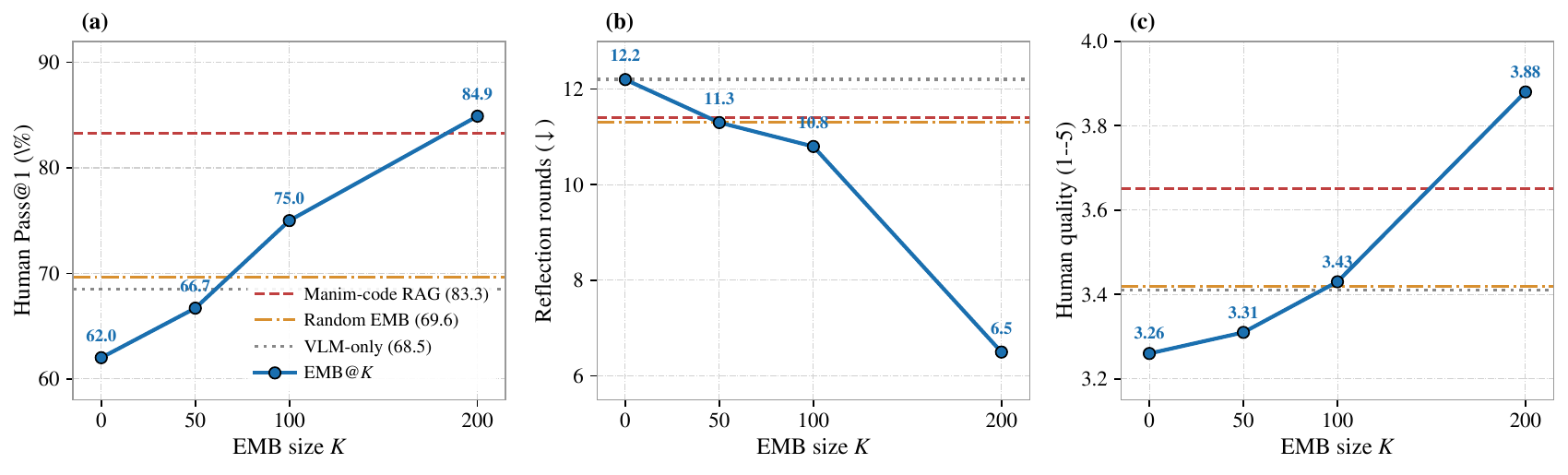}
\caption{\textbf{The descriptive point estimates improve monotonically with EMB size.}
Solid line with markers tracks the frozen EMB snapshot
at $K{\in}\{0,50,100,200\}$ on each of the three headline metrics; horizontal
references are Manim-code RAG (dashed red), Random EMB (dash-dot orange), and
VLM-only (dotted grey). For Human Pass@1 and Human quality, the EMB curve
is above the Manim-code RAG reference at $K{=}200$; for reflection rounds
(lower is better), the corresponding point estimate is substantially lower.}
\label{fig:main-curve}
\end{figure*}

\paragraph{Configurations.}
We compare seven conditions on the same fixed-probe set: VLM reflection only,
an empty EMB (EMB@0), frozen EMB snapshots at $K{\in}\{50,100,200\}$, a
matched-budget Manim-code RAG baseline, and Random EMB. Per-condition
interventions are summarised in Table~\ref{tab:main-results}; full operational
descriptions and the Coder prompt template that turns retrieved records into
the Reference Examples and Known Pitfalls slots are
deferred to Appendix~\ref{sec:appendix-impl-details} and
Appendix~\ref{sec:appendix-coder-prompt}.

\paragraph{Models.}
All text agents and the multimodal VLM reviewer use a single GPT-5.5 family
endpoint; only the reviewer receives rendered keyframes. Decoding settings
are fixed across conditions. The Weak-VLM ablation replaces the reviewer
with Step-3.7-Flash. Renderer version, retrieval
hyperparameters, and per-channel truncation budgets are held constant and
listed in Table~\ref{tab:fixed-hparams}.

\paragraph{Baselines.}
Our main comparison is \textbf{Manim-code RAG}, which extends VLM reflection
with retrieval from a fixed, curated corpus of public Manim documentation and
community examples through the same encoder, index, and truncation budget as
the EMB; the corpus is in-domain, the retrieval machinery is identical, and
it has no exposure to our tasks, so an EMB--RAG gap is evidence for the
self-grown dual-channel content of the EMB rather than for retrieval per se
(Appendix~\ref{sec:appendix-emb-vs-rag}; a documentation-only variant appears
in Table~\ref{tab:ablations}). Comparisons with generic LLM-agent frameworks
and the closest paper-to-animation systems
(Code2Video~\citep{chen2026code2video}, Manimator~\citep{jain2025manimator},
LLM2Manim~\citep{joshi2026llm2manim}, Paper2Video~\citep{zhu2025paper2video})
are discussed in Appendix~\ref{sec:appendix-related-extended}; we do not
benchmark them because their output modality and termination criteria do not
map to a single \textsc{Manim} scene.

\paragraph{Human evaluation.}
54 blind raters supply a binary usability decision and five 1--5 Likert
scores~\citep{likert1932technique}
per first-attempt video across 25 audited fixed-probe tasks. Agreement is low:
Fleiss' $\kappa{=}-0.01$ for binary usability, with $59.4\%$ full
agreement~\citep{fleiss1971measuring}, and the five Likert-dimension ICCs are
0.24, 0.00, 0.10, 0.33, and $-0.09$~\citep{shrout1979intraclass}.
We retain all blind ratings without exclusion, majority voting, or post hoc
adjudication. Recruitment, protocol, and agreement details are in
Appendix~\ref{sec:appendix-human-protocol}.

\paragraph{Metrics.}
We report \textbf{Human Pass@1}, the fraction of first-attempt videos judged by
  blind raters as usable paper-section animations; \textbf{reflection rounds},
  the sum of text-reviewer retries and VLM-driven visual revisions; and
  \textbf{Human quality},
the mean of five 1--5 human rating dimensions: visual design, key-claim
coverage, visual robustness, animation flow, and first-attempt usability. VLM
  scores are auxiliary diagnostics only.

\subsection{Main Results: First-Attempt Generation}
\label{sec:rq1}

For each frozen snapshot, we run the identical fixed-probe set with the same
model, renderer, reflection budget, decoding parameters, and human-scoring
protocol. The fixed-probe design controls for task order, test-time learning,
and prompt-length effects; the Manim-code RAG and Random EMB baselines make the
retrieval comparison stricter. The headline block of Table~\ref{tab:main-results} reports the seven
evaluated conditions and Figure~\ref{fig:main-curve}
visualises the EMB snapshot trends against the three reference baselines.

The descriptive point estimates for EMB@$K$ move monotonically on all three metrics
(Figure~\ref{fig:main-curve}, Table~\ref{tab:main-results}): Pass@1 rises
62.0$\to$84.9, quality 3.26$\to$3.88, and reflection rounds drop
12.2$\to$6.5 from EMB@0 to EMB@200. Relative to Manim-code RAG, EMB@200
has slightly higher point estimates on both human metrics and a substantially
lower reflection-round estimate (6.5 vs.\ 11.4); Random EMB rules out generic
memory-format-context explanations.

\begin{figure*}[!htbp]
  \centering
  \rmfamily\scriptsize
  \setlength{\tabcolsep}{3pt}
  \renewcommand{\arraystretch}{1.0}
  \begin{tabular}{@{}>{\centering\arraybackslash}m{0.08\linewidth}|>{\centering\arraybackslash}m{0.29\linewidth}|>{\centering\arraybackslash}m{0.29\linewidth}|>{\centering\arraybackslash}m{0.29\linewidth}@{}}
    & \textbf{Title scene} & \textbf{Masked LM scene} & \textbf{Fine-tuning scene} \\
    \cmidrule(lr){2-4}
    \shortstack[c]{First\\renderable} &
    \includegraphics[width=\linewidth,height=0.12\textheight,keepaspectratio]{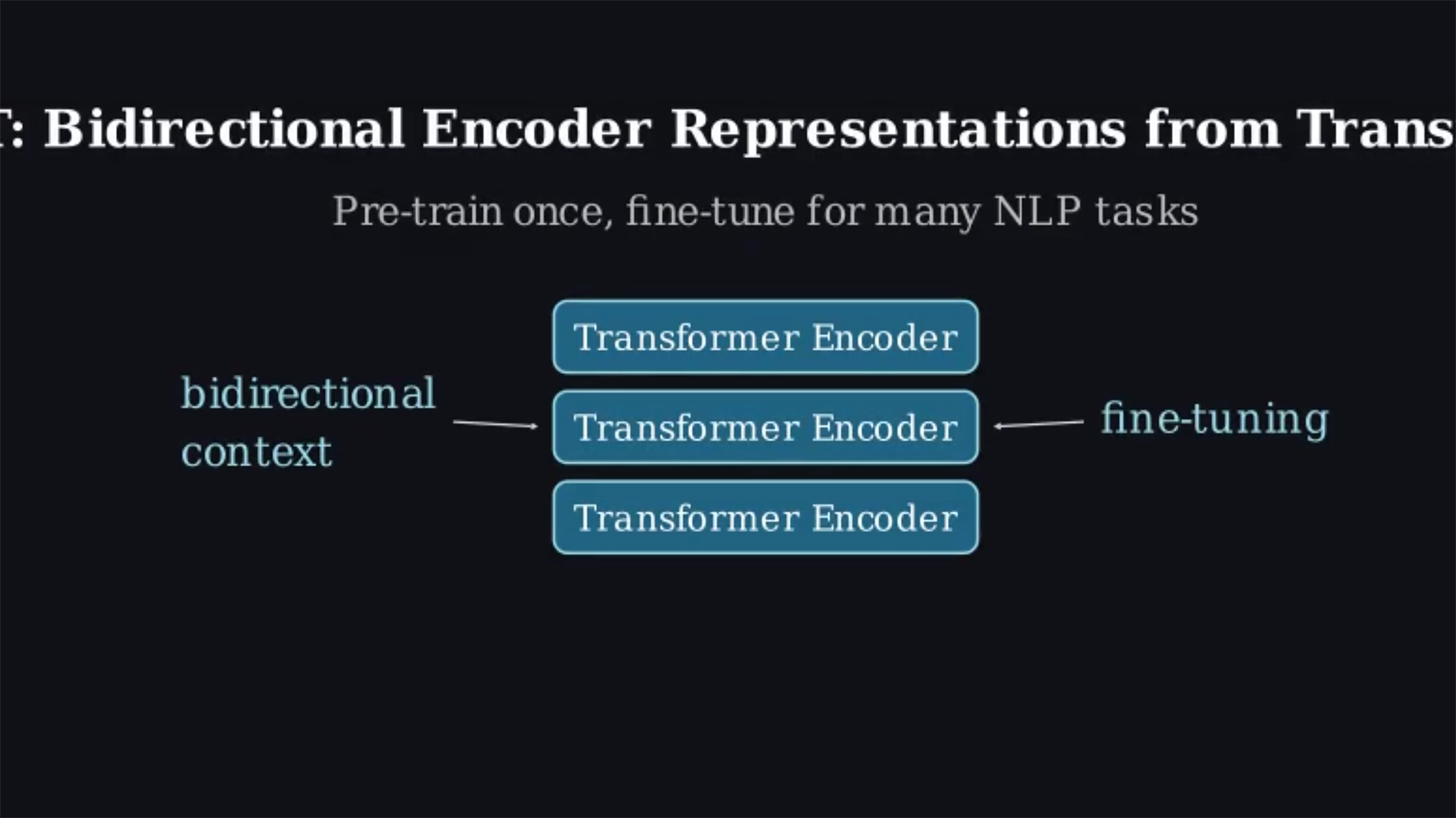} &
    \includegraphics[width=\linewidth,height=0.12\textheight,keepaspectratio]{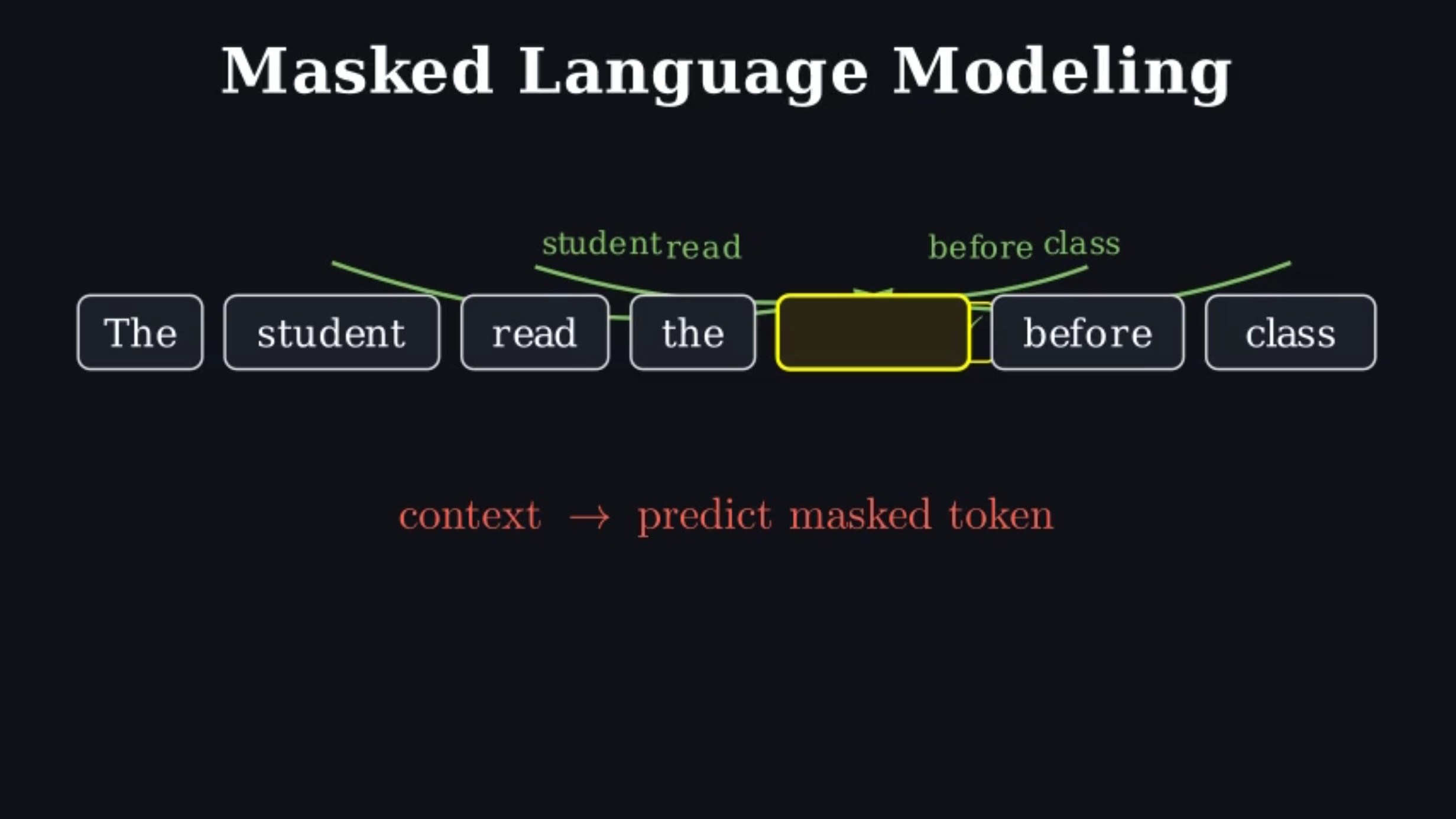} &
    \includegraphics[width=\linewidth,height=0.12\textheight,keepaspectratio]{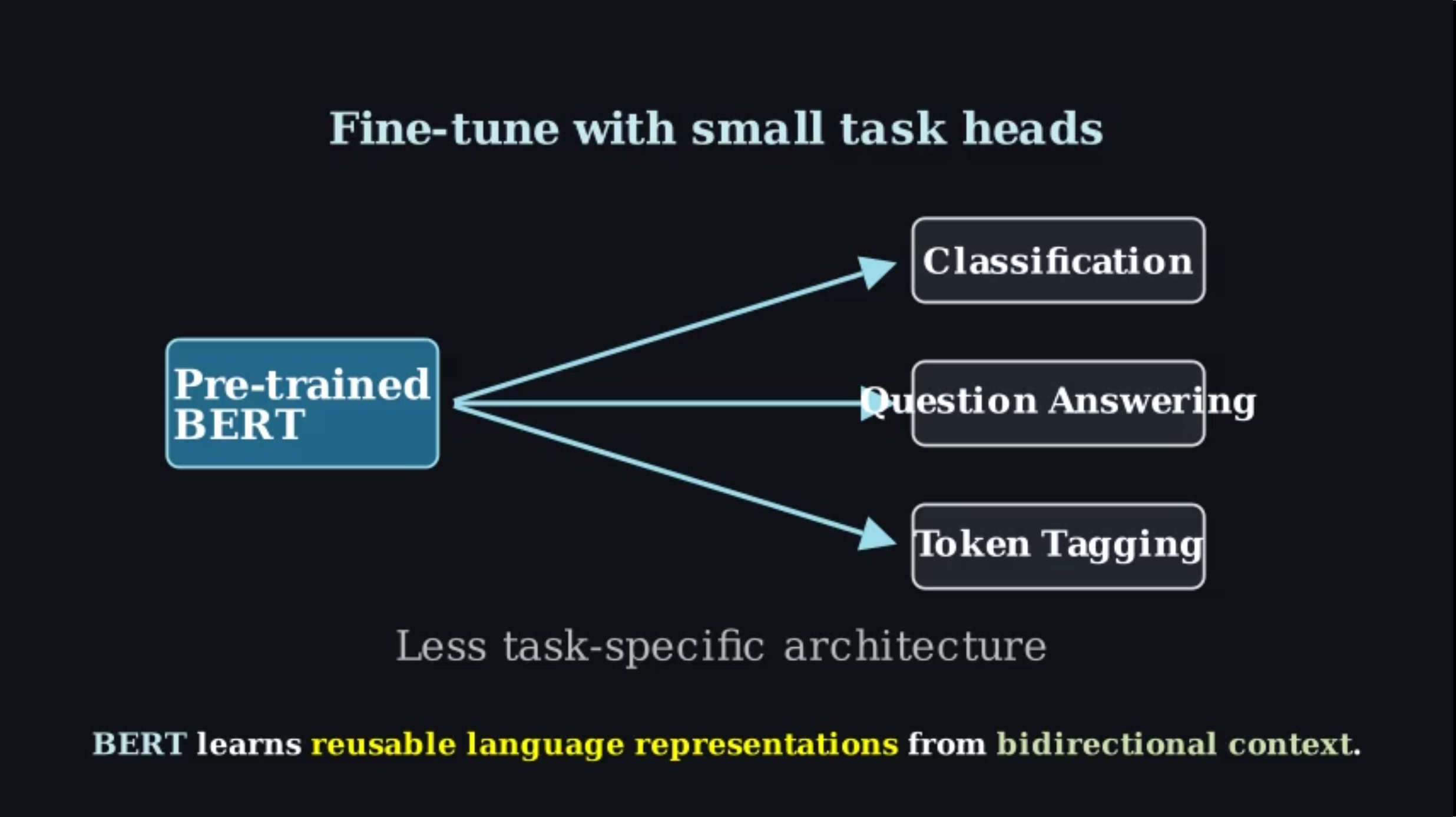} \\
    \cmidrule(lr){2-4}
    \shortstack[c]{Final\\recorded} &
    \includegraphics[width=\linewidth,height=0.12\textheight,keepaspectratio]{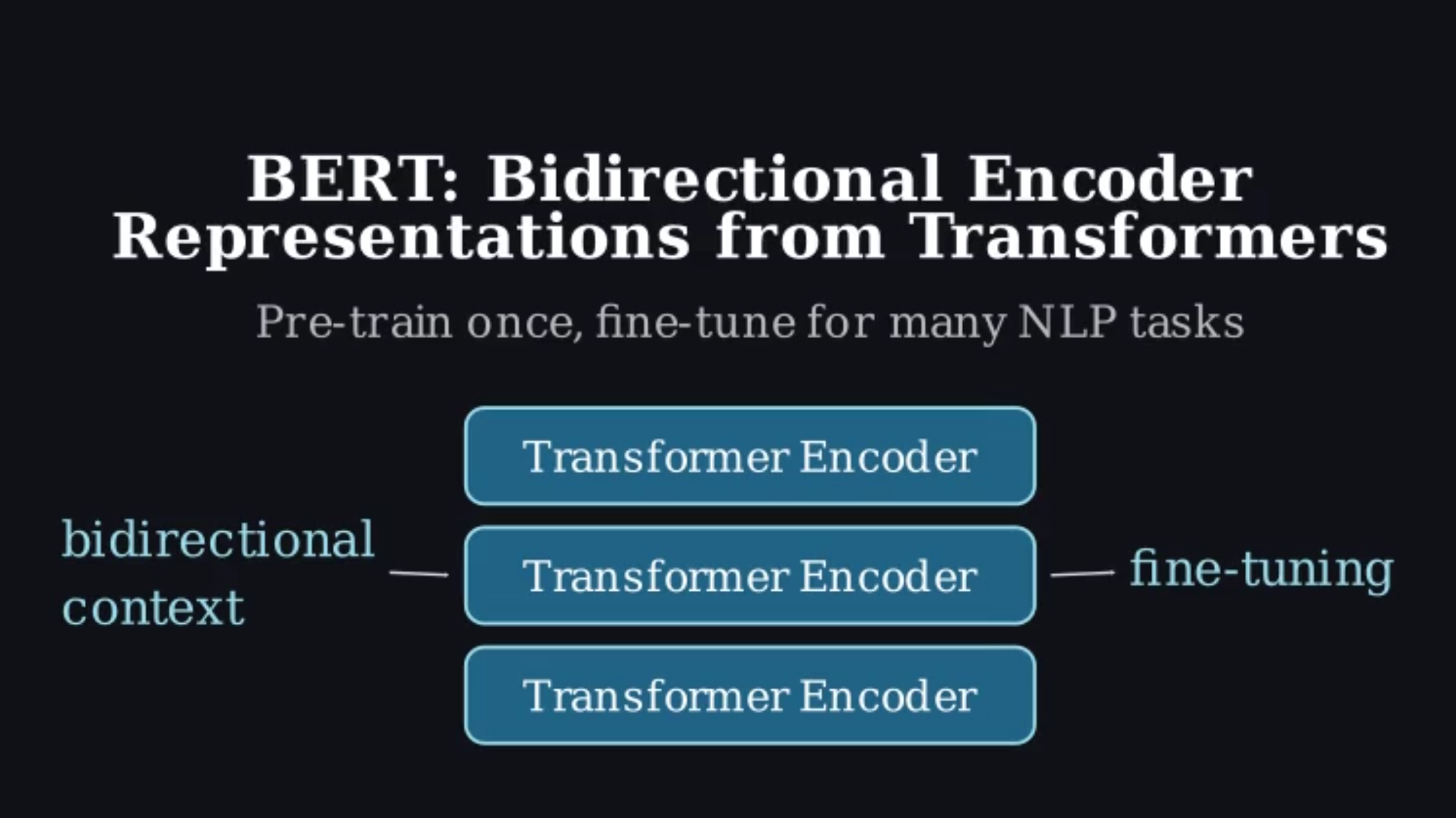} &
    \includegraphics[width=\linewidth,height=0.12\textheight,keepaspectratio]{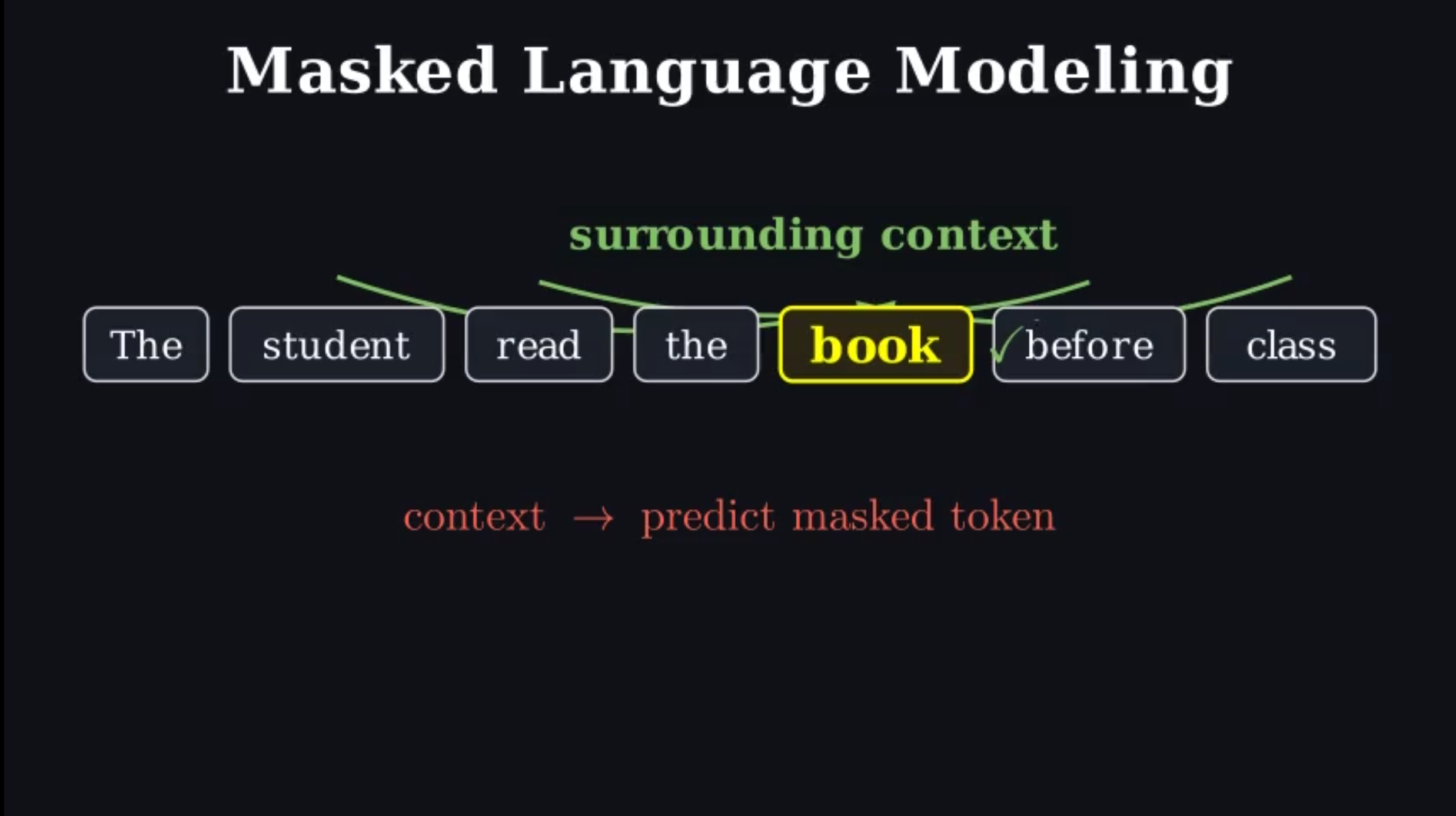} &
    \includegraphics[width=\linewidth,height=0.12\textheight,keepaspectratio]{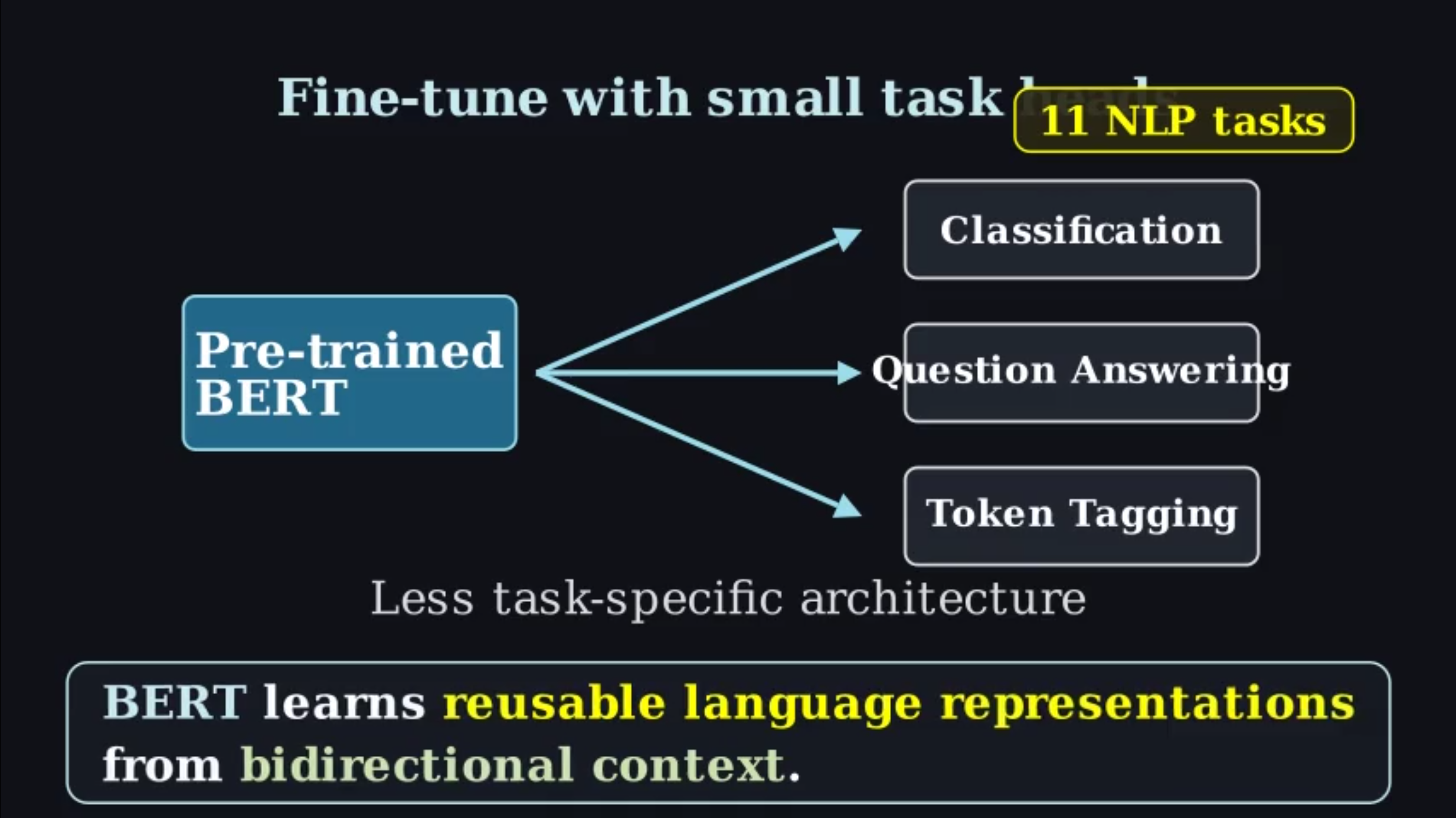} \\
  \end{tabular}
  \caption{\textbf{A compact qualitative example traces three scene revisions for one BERT fixed-probe task.}
  The main paper keeps the qualitative experiment compact by showing three
  representative scene revisions as before/after thumbnails. The title scene
  becomes balanced around the encoder stack, the masked-language-modeling scene
  exposes surrounding context and token prediction, and the fine-tuning scene
  clarifies the downstream-task hierarchy. Appendix~\ref{sec:appendix-qualitative-trace}
  reports the detailed trace outcomes, VLM axes, and larger before/after grid.}
  \label{fig:case-study}
\end{figure*}

Figure~\ref{fig:case-study} gives a compact qualitative example on one BERT
fixed-probe run: visual reflection turns first renderable scenes with weak
layout hierarchy into clearer final frames. Appendix~\ref{sec:appendix-online-curve}
shows the interpolated online curve, and
Appendix~\ref{sec:appendix-qualitative-trace} gives the complete qualitative
trace table and larger before/after montage.

\subsection{Mechanism Analysis}
\label{sec:rq2}

The main results establish that the EMB helps; this section asks which
design choices drive the gain. We run six ablation and control variants on
the same fixed probe (one per row of Table~\ref{tab:ablations}), each
removing or replacing a component of the dual-channel memory or the
consolidation pipeline: channel isolation (dropping $\mathcal{M}^{+}$ or
$\mathcal{M}^{-}$ individually), write-gate removal (keeping every
failed-attempt diagnostic with no minimum-improvement gate), shallow
retrieval ($k_{+}{=}k_{-}{=}1$), rationale-only $\mathcal{M}^{+}$ (dropping
the code excerpt from positive injection), matched-token RAG controls
(documentation-only, with documentation-plus-code represented by the
headline Manim-code RAG row), and a weak-VLM substitution (Step-3.7-Flash
as reviewer). Each variant uses the same human-scoring protocol; the
per-variant mechanism rationale is in
Appendix~\ref{sec:appendix-ablations}.

The channels play differentiated roles: removing $\mathcal{M}^{+}$ collapses
Pass@1 and quality (positive exemplars carry first-attempt benefit), while
removing $\mathcal{M}^{-}$ preserves quality but inflates reflection rounds
(failure memories cut repair cost); the gated full EMB beats ungated
$\mathcal{M}^{-}$ and Top-1 retrieval, limiting context-padding explanations.

\subsection{Reward Reliability, Cost, and Scalability}
\label{sec:rq3}

\begin{figure}[!t]
\centering
\includegraphics[width=\compactfigwidth]{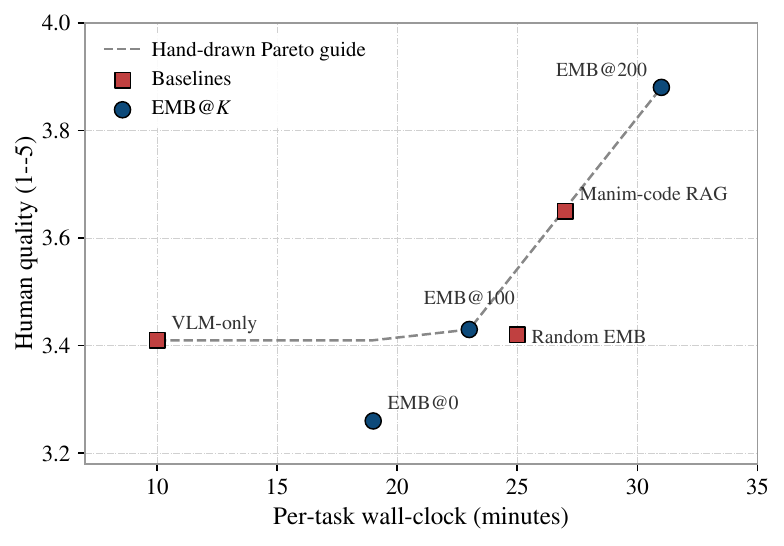}
\caption{\textbf{EMB@200 sits on the cost--quality Pareto frontier.}
Each marker is a condition; $x$ is per-task wall-clock from
Table~\ref{tab:cost}, $y$ is Human quality from
Table~\ref{tab:main-results}. EMB@200 trades 4 additional minutes over
Manim-code RAG for a $+0.23$ gain on Human quality while EMB@0/RAG/Random
cluster around quality $\approx 3.4$. Point estimates from Table~\ref{tab:cost} and Table~\ref{tab:main-results}.}
\label{fig:cost-pareto}
\end{figure}

The VLM that drives revision and gates writes is a noisy internal signal;
VLM--human agreement on the paired probe subset ($n{=}37$ videos) is weak (Pearson
$r{=}-0.17$, Spearman $\rho{=}-0.20$, Cohen's $\kappa{=}-0.07$
~\citep{cohen1960coefficient};
Appendix~\ref{sec:appendix-vlm-human}). Per-task cost is
an estimated 36--38K LLM and 20--23K VLM tokens over 19--31~min, and
Figure~\ref{fig:cost-pareto} shows EMB@200 on the cost--quality Pareto
frontier under a comparable retrieval budget, with higher wall-clock time than Manim-code RAG; per-condition cost and
operational limits are deferred to
Appendix~\ref{sec:appendix-reproducibility} and the Limitations section.

\paragraph{Scaling beyond EMB@200.}
To examine how the system behaves as the memory bank grows beyond the range
used in the main human evaluation, we further evaluate EMB snapshots containing
250, 300, 350, and 400 entries. All four settings use the same 33 fixed-probe
tasks, model configuration, retrieval strategy, and generation pipeline, so the
EMB size remains the primary variable. Table~\ref{tab:emb-scaling} reports token
consumption, reflection rounds, and the internal VLM score for these additional
settings.

\begin{table}[H]
  \centering
  \footnotesize
  \setlength{\tabcolsep}{4pt}
  \renewcommand{\arraystretch}{1.08}
  \begin{tabular}{@{}lccc@{}}
    \toprule
    \textbf{EMB size}
    & \shortstack{\textbf{Token}\\\textbf{consumption}}
    & \shortstack{\textbf{Reflection}\\\textbf{rounds}}
    & \shortstack{\textbf{VLM}\\\textbf{score}} \\
    \midrule
    EMB@250 & 49K & 6.8485 & 86.6500 \\
    EMB@300 & 48K & 6.7273 & 86.2929 \\
    EMB@350 & 48K & 6.5758 & 86.4389 \\
    EMB@400 & 50K & 6.8788 & 86.6202 \\
    \bottomrule
  \end{tabular}
  \caption{\textbf{Scaling diagnostics beyond EMB@200 on the fixed-probe set.}
  The additional settings use the same tasks, model configuration, retrieval
  strategy, and generation pipeline. VLM score is an internal diagnostic; these
  settings were not independently re-evaluated by human raters.}
  \label{tab:emb-scaling}
\end{table}

The results indicate that the scaling trend largely plateaus beyond 250 entries.
Across EMB@250--400, the VLM score remains within a narrow range of
86.29--86.65 and does not improve monotonically as the memory bank expands.
EMB@400 obtains a score of 86.62, close to 86.65 at EMB@250, suggesting that
the small variations across these settings reflect saturation rather than
continued gains. Token consumption remains between 48K and 50K, while the
average number of reflection rounds stays between 6.58 and 6.88.

This stability is consistent with the fixed-depth retrieval design. For each
scene, the system injects only the retrieved top-$k_{+}=2$ positive and
top-$k_{-}=3$ negative memories into the Coder prompt, rather than placing the
entire EMB in the model context. Increasing the total EMB size therefore
expands the candidate pool for vector retrieval without causing the prompt
length or per-task token consumption to grow linearly. The current
implementation uses a fixed-dimensional vector index, whose storage and
retrieval overhead remains small relative to code generation, Manim rendering,
and VLM review within the evaluated range.

We do not repeat human evaluation for these additional settings. Accordingly,
the VLM scores in Table~\ref{tab:emb-scaling} are used only to characterize the
internal scaling trend and should not be interpreted as independent evidence of
human-perceived quality improvements. Overall, larger EMBs remain operationally
stable but provide limited additional benefit at the scale considered here.

\section{Conclusion}
\label{sec:conclusion}

The dual-channel EMB of \textsc{ManimAgent} converts per-task reflection traces into
cross-task experience without weight updates. A fixed-probe snapshot protocol
shows monotonic gains across EMB snapshots, with the mature EMB@200
snapshot matching Manim-code RAG on human metrics while substantially
reducing reflection rounds. Ablations link these gains to the two channels
rather than to retrieval or prompt-length artefacts.

\section*{Limitations}
\paragraph{Evaluation scope.}
\textsc{ManimAgent} is evaluated only with \textsc{Manim} across five
English-language scientific domains; transfer to other renderers or languages,
and effects on student learning, remain~untested.

\paragraph{Scale and evaluator dependence.}
We validate the EMB only through 400 entries and do not extrapolate its stable
trend to production-scale or indefinitely growing memories. Memory maintenance,
robustness to model upgrades, and prompt sensitivity remain open. Because
VLM--human agreement is weak, VLM scores are used only for repair and memory
writing, not as headline quality evidence (Appendices~\ref{sec:appendix-reproducibility}
and~\ref{sec:appendix-vlm-human}).

\paragraph{Baseline and contamination.}
Controls omit long-context retrieval, and split isolation does not remove
within-stream near-duplicates (Appendices~\ref{sec:appendix-emb-vs-rag}
and~\ref{sec:appendix-validator}).

\section*{Ethics and Artifact Release Statement}
Human evaluation involved 54 anonymous, uncompensated volunteers and collected
no direct identifiers or sensitive content. The study was not conducted under
a formal institutional framework. Raw ratings were temporarily stored with
author-only access and deleted within 14 days. The release excludes paper PDFs
and full text and preserves applicable licenses, notices, and source
attribution for code, documentation, videos, and EMB snapshots. Further
recruitment, data-handling, and artifact-release details appear in
Appendices~\ref{sec:appendix-human-protocol}
and~\ref{sec:appendix-dataset-summary}.

\section*{Acknowledgments}
The project is supported in part by research grants from Alberta Machine
Intelligence Institute (Amii); a Canada CIFAR AI Chair, Amii; and the National
Science and Engineering Research Council~(NSERC).

\FloatBarrier
\clearpage
\bibliographystyle{assets/plainnat}
\bibliography{custom}

\clearpage
\appendix
\appendixcontents
\section{Self-Evolving Loop Pseudocode}
\label{sec:appendix-pseudocode}

Algorithm~\ref{alg:self-evolving-loop} states the per-task procedure described in
\S\ref{sec:method-loop}. Symbols carried over from the main text: a task
$\tau=(s,r,d)$ comprises section text $s$, scene role
$r\in\{\textsc{background},\allowbreak \textsc{method},\allowbreak \textsc{experiment},\allowbreak \textsc{conclusion}\}$,
and domain tag $d$ (\S\ref{sec:method-overview}); the two retrieval sizes are pinned at
$k_{+}=2$ and $k_{-}=3$ (\S\ref{sec:method-emb}). The render \texttt{result} returned by
$R$ exposes a fault \texttt{category} and a Python \texttt{traceback}, both consumed by
\textsc{LessonDistiller} when distilling text-channel pitfalls. \textsc{VLM} additionally
emits rubric \texttt{axes} and per-axis \texttt{issues}; these are persisted to the run
log for analysis but are not read elsewhere by the loop.

\begin{algorithm*}[t]
\small
\caption{The self-evolving loop combines per-scene text and visual reflection, best-of-$N$ delivery, and two distillation paths within a dual-channel EMB.}
\label{alg:self-evolving-loop}
\begin{algorithmic}[1]
\Require Task stream $(\tau_1,\tau_2,\ldots)$; agents $\Sigma$ (Storyboarder), Coder,
Reviewer, VLM, VisualReviser, RationaleWriter, LessonDistiller; Renderer $R$ with static
check; sentence encoder $\phi$; thresholds $\theta_{\mathrm{conv}}$, $\theta_{\mathrm{high}}$,
$\Delta_{\mathrm{fail}}$; budgets $T_{\mathrm{text}}$, $T_{\mathrm{vis}}$.
\State $\mathcal{M}\leftarrow\mathcal{M}^{+}\cup\mathcal{M}^{-}\leftarrow\emptyset$
\For{each task $\tau=(s,r,d)$ in the stream}
    \State $\{\sigma_i\}_{i=1}^{n}\leftarrow\Sigma(\tau)$ \Comment{decompose into scenes}
    \For{each scene $\sigma_i$}
        \State $e_i\leftarrow\phi([s;r])$;\quad $E^{+}\leftarrow\mathrm{top}\text{-}k_{+}(\mathcal{M}^{+},e_i)$;\quad $E^{-}\leftarrow\mathrm{top}\text{-}k_{-}(\mathcal{M}^{-},e_i)$
        \State $c\leftarrow\mathrm{Coder}(\sigma_i,E^{+},E^{-})$;\quad $\mathrm{textTrace}\leftarrow[]$
        \For{$t_{\mathrm{tx}}=0,\ldots,T_{\mathrm{text}}$} \Comment{text-reflection: crash $\to$ retry}
            \State $\mathrm{result}\leftarrow R(c)$;\quad append $(c,\mathrm{result})$ to $\mathrm{textTrace}$ \Comment{static check, then sandbox render}
            \If{$\mathrm{result}$ is success} \State \textbf{break} \EndIf
            \State $(\mathrm{hint},\mathrm{dec})\leftarrow\mathrm{Reviewer}(\mathrm{result})$
            \If{$\mathrm{dec}=\texttt{give\_up}$ \textbf{or} same error category twice} \State \textbf{break} \EndIf
            \State $c\leftarrow\mathrm{Coder}(\sigma_i,E^{+},E^{-},\mathrm{hint},\mathrm{result})$ \Comment{regenerate from scratch}
        \EndFor
        \If{$\mathrm{result}$ is not success} \State \textbf{continue} \Comment{scene skipped; no usable transition exists}
        \EndIf
        \State $\mathrm{cands}\leftarrow[]$;\quad $\mathrm{visTrace}\leftarrow[]$
        \For{$t_{\mathrm{vs}}=0,\ldots,T_{\mathrm{vis}}$} \Comment{visual-reflection: renderable $\to$ revise}
            \State $K\leftarrow R.\mathrm{sample\_keyframes}(c,n_{K}=4)$
            \State $(\mathrm{axes},u,\mathrm{issues},\mathrm{instr},\mathrm{dec})\leftarrow\mathrm{VLM}(s,\sigma_i,K)$
            \State append $(c,u,\mathrm{instr})$ to $\mathrm{visTrace}$;\quad append $(c,u)$ to $\mathrm{cands}$
            \If{$u\geq\theta_{\mathrm{conv}}$} \State $\mathrm{dec}\leftarrow\texttt{pass}$ \Comment{auto-pass override} \EndIf
            \If{$\mathrm{dec}\in\{\texttt{pass},\texttt{fail}\}$} \State \textbf{break} \EndIf
            \State $c\leftarrow\mathrm{VisualReviser}(c,\mathrm{instr})$ \Comment{not the Coder}
        \EndFor
        \State $(c^{\star},u^{\ast})\leftarrow\arg\max_{(c,u)\in\mathrm{cands}}u$ \Comment{best-of-$N$}
        \If{$u^{\ast}\geq\theta_{\mathrm{high}}$}
            \State $\rho\leftarrow\mathrm{RationaleWriter}(s,\sigma_i,c^{\star})$
            \State $\mathcal{M}^{+}\leftarrow\mathcal{M}^{+}\cup\{(e_i,\sigma_i,\rho,c^{\star},u^{\ast})\}$
        \EndIf
        \For{each text transition $(c^{(t-1)},c^{(t)})$ in $\mathrm{textTrace}$ with $c^{(t-1)}{:}\,\mathrm{err}$, $c^{(t)}{:}\,\mathrm{ok}$}
            \State $\ell\leftarrow\mathrm{LessonDistiller}(c^{(t-1)},c^{(t)},\mathrm{category},\mathrm{traceback})$
            \State $\mathcal{M}^{-}\leftarrow\mathcal{M}^{-}\cup\{(e_i,\sigma_i,\ell)\}$
        \EndFor
        \For{each visual transition $(c^{(t-1)},c^{(t)})$ in $\mathrm{visTrace}$ with $u(t)-u(t-1)\geq\Delta_{\mathrm{fail}}$}
            \State $\ell\leftarrow\mathrm{LessonDistiller}(c^{(t-1)},c^{(t)},\mathrm{instr}^{(t-1)})$
            \State $\mathcal{M}^{-}\leftarrow\mathcal{M}^{-}\cup\{(e_i,\sigma_i,\ell)\}$
        \EndFor
    \EndFor
\EndFor
\State \Return $\mathcal{M}$
\end{algorithmic}
\end{algorithm*}

\section{Implementation Details}
\label{sec:appendix-impl-details}

This appendix collects per-condition descriptions for the experimental setup
in \S\ref{sec:exp-setup}, together with renderer-sandbox configuration, length
caps, and short-circuit rules referenced from \S\ref{sec:method-loop} and
\S\ref{sec:method-emb} but kept out of the body for space. All numbers are
pinned in run manifests.

\paragraph{Per-condition descriptions.}
\textbf{VLM reflection only} uses the same renderer, VLM reviewer, and
visual-revision loop as \textsc{ManimAgent} but disables retrieval and memory
writing. \textbf{EMB@0} keeps the same prompt structure with empty
Reference Examples and Known Pitfalls slots, isolating
prompt-structure effects from retrieved content. \textbf{EMB@$K$} retrieves
from a frozen self-grown EMB snapshot containing $K$ consolidated records,
each produced by the best-of-$N$ delivery rule of \S\ref{sec:method-loop} and
the LLM distillation rules of \S\ref{sec:method-emb}. \textbf{Random EMB}
preserves record count, format, and token budget but shuffles records across
  tasks, breaking content--query~\mbox{alignment.}

  \paragraph{Reflection-round accounting.}
  For task $\tau$, we define
  $\mathrm{RR}(\tau)=\sum_{i\in\tau}(n^{\mathrm{retry}}_{i,\mathrm{text}}+
  n^{\mathrm{revision}}_{i,\mathrm{visual}})$. The first term counts retries
  actually performed after a failed static check or render; the second counts
  revisions actually performed in response to VLM feedback after a successful
  render. Initial code generation is excluded. We report the mean of
  $\mathrm{RR}(\tau)$ over all 33 fixed-probe tasks. A scene skipped before a
  successful render contributes only its executed text retries and zero visual
  revisions.

\paragraph{Renderer sandbox and static checker.}
The \textsc{Renderer} executes each candidate as a subprocess under POSIX
\texttt{rlimit} (wall $180$\,s, CPU $120$\,s, RAM $4$\,GB) with the Cairo
backend. Before invocation, the static checker rejects forbidden imports
(\texttt{os}, \texttt{subprocess}, \texttt{socket}, \ldots), forbidden calls
(\texttt{open}, \texttt{eval}, \texttt{exec}), and missing structure (the script
must declare a \texttt{Scene} subclass matching $\sigma_i.\mathrm{name}$ and
call \texttt{self.play} at least twice). On failure the renderer returns a
typed result whose category is \texttt{python}, \texttt{latex},
\texttt{manim\_runtime}, \texttt{timeout}, or \texttt{unknown}, together with a
traceback tail and any TeX log excerpt.

The text \textsc{Reviewer} produces $\le 60$-word hints; the \textsc{LessonDistiller} is
capped at $\le 1500$ output~tokens.

\paragraph{$\mathcal{M}^{-}$ field length caps.}
Trigger $\le 400$ chars, root cause $\le 400$, fix recipe $\le 400$, code
fragments $\le 800$ chars each, free-form diagnostic $\le 1000$ chars. The
\textsc{RationaleWriter} output stored in $\mathcal{M}^{+}$ is capped at
$\le 400$ chars. Retrieval-time truncations match Table~\ref{tab:fixed-hparams}.

\paragraph{Short-circuit and tie-break rules.}
The text-reflection loop exits when $t{=}T_{\mathrm{text}}{=}2$, on the
\textsc{Reviewer}'s \texttt{give\_up} decision, or when the same error
category recurs twice (preventing reflection from looping inside the same kind
of bug). The visual-reflection loop exits when $t{=}T_{\mathrm{vis}}{=}2$ or on
\texttt{pass}/\texttt{fail}; an auto-pass override fires when
$u(t)\ge\theta_{\mathrm{conv}}=90$ regardless of the verdict of the VLM.
Best-of-$N$ delivery ships
$c^{\star}=\arg\max_{t}u(t)$ across renderable candidates; ties are broken by
earliest $t$. These rules are enacted by
Algorithm~\ref{alg:self-evolving-loop} in
Appendix~\ref{sec:appendix-pseudocode}.

\section{Extended Related Work}
\label{sec:appendix-related-extended}

This section extends \S\ref{sec:related-work} with programmatic animation,
reflection-based agents, and parameter-free inference-time optimization.

\paragraph{Programmatic animation and visual education.}
Programmatic animation systems use executable graphics code as an intermediate
representation for visual explanations. \textsc{Manim} provides the rendering substrate for
mathematical animations~\citep{The_Manim_Community_Developers_Manim_Mathematical_2026}, and recent LLM-based systems have begun
to generate such animations or videos automatically. Code2Video generates educational videos
from code~\citep{chen2026code2video}; Manimator converts research papers into visual
explanations~\citep{jain2025manimator}; LLM2Manim studies pedagogy-aware generation of
STEM animations~\citep{joshi2026llm2manim}; and Paper2Video broadens the setting to video
generation from scientific papers~\citep{zhu2025paper2video}. These systems establish the
feasibility of paper- or code-conditioned visual explanation, but they primarily treat each
input as an isolated generation problem. \textsc{ManimAgent} targets the same output medium,
but asks whether the agent can retain transferable visual-programming lessons across tasks.

\paragraph{Reflection and parameter-free optimization.}
Self-Refine~\citep{NEURIPS2023_91edff07}, Reflexion~\citep{NEURIPS2023_1b44b878}, and
CRITIC~\citep{ICLR2024_fef12656} revise from feedback within a task, while ReAct
connects reasoning with tool feedback~\citep{yao2023react}. GEPA
~\citep{ICLR2026_0e9e708b} and Training-Free
GRPO~\citep{cai2025trainingfreegrpo} optimize textual control signals from
rollouts without weight updates. \textsc{ManimAgent} follows the intra-task
reflection pattern but uses EMB to distil positive and negative experiences
into a frozen cross-task store retrieved on disjoint probes. These directions
are complementary rather than matched retrieval baselines; combining them is
future work.

\section{Comparison with Prior Systems}
\label{sec:appendix-comparison}

Table~\ref{tab:comparison} contrasts \textsc{ManimAgent} with prior animation-from-paper
pipelines (Manimator, Code2Video, LLM2Manim) and the closest lifelong-learning
agent (Voyager) along four axes: reflection style, VLM-based judging, cross-task memory, and
fixed-probe evaluation. The prior animation pipelines reflect within a task but never carry
information across tasks; Voyager carries cross-task memory as a single library of successful
executable skills. \textsc{ManimAgent} is the only system that combines
multi-agent visual reflection, a structured multi-axis VLM judge, a self-grown dual-channel
memory, and a fixed-probe snapshot evaluation as the primary metric.

\begin{table*}[t]
  \centering
  \small
  \setlength{\tabcolsep}{4pt}
  \resizebox{\linewidth}{!}{%
  \begin{tabular}{@{} l c c c c @{}}
    \toprule
    \textbf{System} & \textbf{Reflection} & \textbf{VLM judge} & \textbf{Cross-task memory} & \textbf{Fixed-probe test} \\
    \midrule
    Manimator~\citep{jain2025manimator}         & -- & -- & -- & -- \\
    Code2Video~\citep{chen2026code2video}        & multi-agent & VLM critic & -- & -- \\
    LLM2Manim~\citep{joshi2026llm2manim}         & compile-error & -- & -- & -- \\
    Voyager~\citep{wang2023voyager}               & --           & -- & self-grown executable skills & partial \\
    \textbf{\textsc{ManimAgent}} & multi-agent, visual & structured, multi-axis & \textbf{dual-channel, self-grown} & \textbf{primary metric} \\
    \bottomrule
  \end{tabular}%
  }
  \caption{\textbf{\textsc{ManimAgent} combines self-grown dual-channel memory with fixed-probe
  evaluation.} Prior systems either lack cross-task memory, retain a single channel of successful
  executable skills, or do not evaluate with held-out snapshots.}
  \label{tab:comparison}
\end{table*}

\section{EMB vs. RAG and Retrieval Terminology}
\label{sec:appendix-emb-vs-rag}

The EMB uses retrieval, but it is not standard retrieval-augmented generation (RAG) with a
different name. Table~\ref{tab:emb-vs-rag} contrasts the two on five axes: corpus, growth,
write rule, content, and prompt role. The corpus of the EMB is self-generated task traces; it
grows after each converged task under VLM-gated write rules; it stores both success
  rationales and VLM-screened failure patterns; and it injects them into two distinct prompt
slots. A static Manim-code RAG corpus can still strengthen first drafts, which is why
\S\ref{sec:exp-setup} includes it as a baseline; the fixed-probe experiment then tests
whether mature EMB snapshots outperform that static corpus under matched token budget,
prompt length, retrieval depth, encoder, and index, so an EMB--RAG gap is evidence for
trajectory-derived cross-task memory rather than retrieval alone.

\begin{table}[t]
  \centering
  \small
  \setlength{\tabcolsep}{3pt}
  \begin{tabular}{p{0.22\linewidth}p{0.32\linewidth}p{0.32\linewidth}}
    \toprule
    \textbf{Axis} & \textbf{Standard RAG} & \textbf{EMB (ours)} \\
    \midrule
    Corpus & fixed docs/tutorials/code & self-generated task traces \\
    Growth & static during evaluation & grows after converged tasks \\
    Write rule & none & VLM-gated success and failure writes \\
    Content & human-written references & success rationales plus VLM-screened failure patterns \\
    Prompt role & one context slot & Reference Examples and Known Pitfalls \\
    \bottomrule
  \end{tabular}
  \caption{\textbf{EMB differs from static RAG in source, growth, write rule,
  content, and prompt role.}}
  \label{tab:emb-vs-rag}
\end{table}

\paragraph{Retrieval terminology.}
The EMB lookup is implemented as a retrieval layer over a Faiss IndexFlatIP. We call this a
retrieval layer in the codebase, an engineering term for the nearest-neighbour lookup
component, and it should not be confused with the static Manim-code RAG baseline, which
retrieves from a fixed external corpus rather than from a self-grown, polarity-tagged memory
bank. Both use the same encoder (all-MiniLM-L6-v2, 384-dimensional embeddings) and the same
index type (Faiss IndexFlatIP with cosine similarity), so any retrieval-quality gap reflects
differences in retrieved content rather than retrieval machinery.

\paragraph{Long-context baseline scope.}
The matched-budget controls do not retrieve raw historical code, complete
reflection trajectories, or reviewer feedback. A fair long-context comparison
would need to hold the source tasks, retriever, candidate count, and final token
budget fixed while varying only whether historical information is injected as
raw trajectories or distilled positive and negative experiences. We leave this
comparison to future work.

\section{Dataset Validation}
\label{sec:appendix-validator}

The released dataset is accompanied by an automated validator. It enforces zero paper
overlap across the memory-building, fixed-probe, and cross-test splits; zero
overlap between model-input and evaluation-only fields; valid source-text
hydration for all headline tasks; distribution match across split strata; contamination
and license metadata; snapshot plausibility for
$K\in\{0,50,100,200\}$; and the release policy excluding raw full-paper text
and embedded human scores.

\noindent Strict paper-readiness validation additionally requires the
human-scores file to be non-empty and should be run after output-level
annotations and manifests have been collected. The dataset is release-ready
before this final result-validation~pass.

\paragraph{Within-stream duplication.}
Paper-level isolation prevents overlap between the memory-building and probe
sets but does not rule out semantically near-duplicate tasks within the
memory-building stream. The current validator records contamination metadata
but does not perform paraphrase-level deduplication; this remains future work.

\section{Released Dataset Summary}
\label{sec:appendix-dataset-summary}

\begin{figure}[t]
\centering
\includegraphics[width=\compactfigwidth]{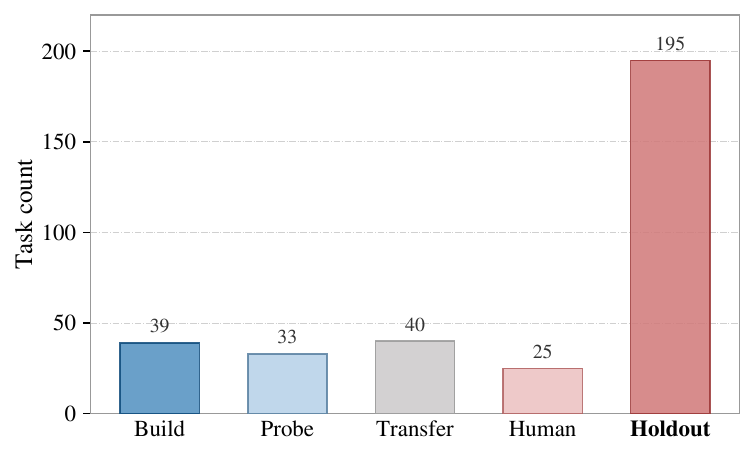}
\caption{\textbf{Dataset split sizes are deliberately small for the three reported subsets.} The memory-building, fixed-probe,
cross-test, and human-scoring subsets support construction and evaluation;
the quarantined holdout is excluded from headline claims.}
\label{fig:dataset-splits}
\end{figure}

Table~\ref{tab:dataset-summary} summarises the released artefacts, and
Figure~\ref{fig:dataset-splits} visualises the relative split sizes: the three
headline subsets (memory-building 39, fixed-probe 33, cross-test 40) are
deliberately small relative to the quarantined holdout of 195 papers, which is
excluded from every reported number. The release includes dataset metadata,
headline tasks, quarantined holdout tasks, and paper metadata. Raw full-paper
text, draft annotations, example experiment manifests, and output-level human
scores are excluded from the model-visible release.
The release manifest states licensing terms for system-generated code, videos,
and EMB snapshots and preserves original notices and source attribution for
Manim documentation, tutorials, and code~\mbox{examples.}

\begin{table*}[t]
\centering
\small
\setlength{\tabcolsep}{4pt}
\begin{tabular}{p{0.15\linewidth}p{0.08\linewidth}p{0.32\linewidth}p{0.32\linewidth}}
\toprule
Split & Size & Released fields & Role \\
\midrule
Memory-building split & 39 & Identifiers, target-unit anchors, metadata, scene roles &
Grows the EMB before snapshot freezing \\
Fixed-probe split & 33 & Identifiers, target-unit anchors, metadata, evaluation-only
rubrics & Evaluates every frozen EMB snapshot in read-only mode \\
Cross-test split & 40 & Identifiers, target-unit anchors, domain tags, metadata &
Exploratory transfer probe \\
Human-scoring subset & 25 & Candidate task identifiers; output-level ratings are
stored separately after annotation & Blind Human Pass@1
and Human quality \\
Quarantined holdout & 195 & debug metadata only (177 fallback, 18 promoted) &
Excluded from all headline claims \\
\bottomrule
\end{tabular}
\caption{\textbf{The released dataset separates construction, evaluation, and exploratory transfer.} The benchmark separates memory construction,
fixed-snapshot evaluation, and exploratory transfer testing; human scores remain
separate from model-visible inputs.}
\label{tab:dataset-summary}
\end{table*}

\subsection{Representative Fixed-Probe Examples}
\label{sec:appendix-dataset-examples}

Table~\ref{tab:dataset-examples} gives representative records sampled from the
human-evaluation candidate pool in the fixed-probe split. The
selection covers all five dataset domains, all four scene roles, and all three
difficulty levels. At inference time the system receives the paper title,
target-unit text and anchor, domain, and scene role in the model input;
human ratings and output annotations remain evaluation-only.
Table~\ref{tab:dataset-examples} lists representative fixed-probe tasks for
coverage; Table~\ref{tab:dataset-record-example} shows a schema-complete
example record.

\begin{table*}[t]
\centering
\small
\setlength{\tabcolsep}{4pt}
\begin{tabular}{p{0.24\linewidth}p{0.11\linewidth}p{0.14\linewidth}p{0.09\linewidth}p{0.32\linewidth}}
\toprule
Task identifier & Domain & Scene role & Difficulty & Paper / target-unit anchor \\
\midrule
\shortstack[l]{\texttt{2305\_18290\_}\\[-1pt]\texttt{section\_experiments}} & CS & Experiment & Easy &
Direct Preference Optimization / Experiments \\
\shortstack[l]{\texttt{1609\_07132\_}\\[-1pt]\texttt{section\_method}} & Math & Method & Medium &
Optimization Methods for Large-Scale Machine Learning / Method \\
\shortstack[l]{\texttt{1207\_7235\_}\\[-1pt]\texttt{section\_results}} & Physics & Experiment & Hard &
Quantum Entanglement and Topological Order / Results \\
\shortstack[l]{\texttt{quant-ph\_0508027\_}\\[-1pt]\texttt{section\_introduction}} & Quantum & Background & Easy &
A One Way Quantum Computer / Introduction \\
\shortstack[l]{\texttt{1706\_01427\_}\\[-1pt]\texttt{section\_discussion}} & Economics & Conclusion & Hard &
Machine Learning: An Applied Econometric Approach / Discussion \\
\bottomrule
\end{tabular}
\caption{\textbf{Five representative fixed-probe tasks anchor the human-evaluation pool.}
The examples are taken from the released benchmark and illustrate
domain, scene-role, and difficulty coverage without reproducing full paper~text.}
\label{tab:dataset-examples}
\end{table*}

\subsection{Example Dataset Record}
\label{sec:appendix-dataset-record}

Table~\ref{tab:dataset-record-example} gives an illustrative, shortened record
whose field groups match the benchmark and private evaluation-sidecar schema. Long section excerpts
are abbreviated and no raw full-paper text is reproduced. Evaluation-only
annotations are structurally shown to make isolation explicit; they are never
presented to the generation system or written to the EMB.

\begin{table*}[t]
\centering
\footnotesize
\setlength{\tabcolsep}{4pt}
\begin{tabular}{p{0.19\linewidth}p{0.22\linewidth}p{0.50\linewidth}}
\toprule
Visibility group & Field & Illustrative shortened value \\
\midrule
Model-visible input
& Paper title & \emph{Attention Is All You Need} \\
& Abstract excerpt & Abstract field present; shortened excerpt where available \\
& Target-unit anchor & Introduction \\
& Target-unit excerpt & ``Recurrent models typically factor computation along symbol positions \ldots'' \\
& Required prior context & Not required \\
& Prior objects & None \\
& Scene role & Background \\
& Domain & Computer science \\
& Source type & Section \\
\midrule
Evaluation-only fields
& Key claims & Human-audited local claims in a private evaluation sidecar \\
& Reference scene plan & Expected pedagogical scene progression \\
& Task-specific rubric & Alignment, robustness, coverage, flow, and usability criteria \\
& VLM diagnostic rubric & Logic-flow, layout/occlusion, and accuracy checks \\
& Condition identifiers & Hidden system condition and snapshot identifier \\
& Fatal issue flags & Closed-set annotation options used by blind raters \\
\midrule
Metadata / release
& Split & Fixed-probe split \\
& Paper / section identifier & \texttt{1706.03762} / Introduction \\
& Licence metadata & arXiv non-exclusive distribution licence; recorded source URL \\
& Contamination metadata & Pre-cutoff publication stratum; cutoff reference 2023-10-01 \\
& Quarantine metadata & Hydration status valid; included in headline probe \\
\midrule
Human outputs
& Ratings sidecar & Human Pass@1 votes, Likert ratings, flags, and optional notes are
stored separately from model-visible records. \\
\bottomrule
\end{tabular}
\caption{\textbf{This illustrative benchmark record covers the full schema.} The example
shows model-visible inputs, evaluation-only fields, and release metadata while
shortening textual content and isolating output-level human~\mbox{annotations.}}
\label{tab:dataset-record-example}
\end{table*}

\section{Human Annotation Protocol}
\label{sec:appendix-human-protocol}

The main quality metrics are based on blind human scoring of first-attempt
videos. Human scores are stored separately from the model-visible task file and
are never embedded in task records or shown to any system. The release layout
deliberately keeps collected output-level annotations outside the model-visible
task collection.

\paragraph{Raters.}
54 raters take part in the blind human evaluation, and each evaluated video
receives two or three independent ratings. Raters are recruited from graduate programmes
in computer science, applied mathematics, physics, and economics. No rater has prior
exposure to \textsc{Manim} code generation or to the specific papers in the evaluation
stream. 77\% identify CS/AI as their primary discipline, 11\% economics/social science,
5\% mathematics/statistics, and 5\% physics/quantum. 42\% report frequently reading
research papers; 64\% have occasional prior experience evaluating technical diagrams or
animations, while 33\% have regular experience. Before main annotation, each rater
completes a one-hour calibration session on 5 held-out calibration examples covering the
full quality~range.

\paragraph{Ethics and data retention.}
Raters are anonymous, uncompensated volunteers. We collect no names, contact
information, or other direct identifiers. They assess only non-sensitive
scientific animations. Raw ratings are stored temporarily on an author-only
rented server and deleted within 14 days. The study is not conducted under a
formal institutional framework.

\paragraph{Background questionnaire.}
Before rating any videos, each rater completes a three-item background form:
(i) primary discipline (computer science/AI, mathematics/statistics,
physics/quantum, economics/social science, or other/no specific background);
(ii) frequency of reading research papers or technical literature (rarely/never,
occasionally, or frequently); and (iii) prior experience evaluating technical
diagrams, animations, or other visualisations (none, occasional viewing/evaluation,
or frequent creation/use/evaluation). These responses characterize the
rater pool and are not used as model inputs.

\paragraph{Inputs shown to raters.}
For each sample, raters see the paper title, target-unit title and text, scene
role, key claims, reference scene plan, task-specific rubric, and generated
first-attempt video. The reference scene plan is guidance, not a required script.
Raters do not see the system condition, EMB size, VLM score, prompt, code,
revision trace, or any filename that reveals the condition.

\paragraph{Sampling plan.}
Human-scoring tasks are sampled from successfully hydrated fixed-probe tasks
after target annotations have been human-audited.
We sample 25 tasks, stratified by scene role (\textsc{background},
\textsc{method}, \textsc{experiment}, \textsc{conclusion}), domain (five
domains), and difficulty (three bins: number of distinct visual elements, number
of animation stages, presence of equations). Each video receives three
independent ratings. The main-table plan covers 7 conditions: VLM reflection
only, Manim-code RAG, Random EMB, EMB@0, EMB@50, EMB@100, and EMB@200.
Applied to 25 tasks, this yields 175 evaluated videos. Raters may
evaluate videos from multiple conditions. Output-level ratings for
all conditions are intentionally kept outside the model-visible dataset.

\paragraph{Questionnaire and rating dimensions.}
For each video, the form first asks the binary Human Pass@1 question, then five
1--5 Likert items (1=very poor, 3=average, 5=very good):

\begin{itemize}
\item \textbf{Visual design.} Are composition, colour, information
  hierarchy, and overall visual presentation clear and coherent?
\item \textbf{Key-claim coverage.} Are the central claims visually
  expressed?
\item \textbf{Visual robustness.} Are elements readable, free of
  overlap, and correctly rendered?
\item \textbf{Animation flow.} Is the pacing, ordering, and transition
  between scenes appropriate?
\item \textbf{First-attempt usability.} Could this video be used as a
  paper-section animation without major structural revision?
\end{itemize}

\begin{figure*}[!t]
\centering
\includegraphics[width=\linewidth]{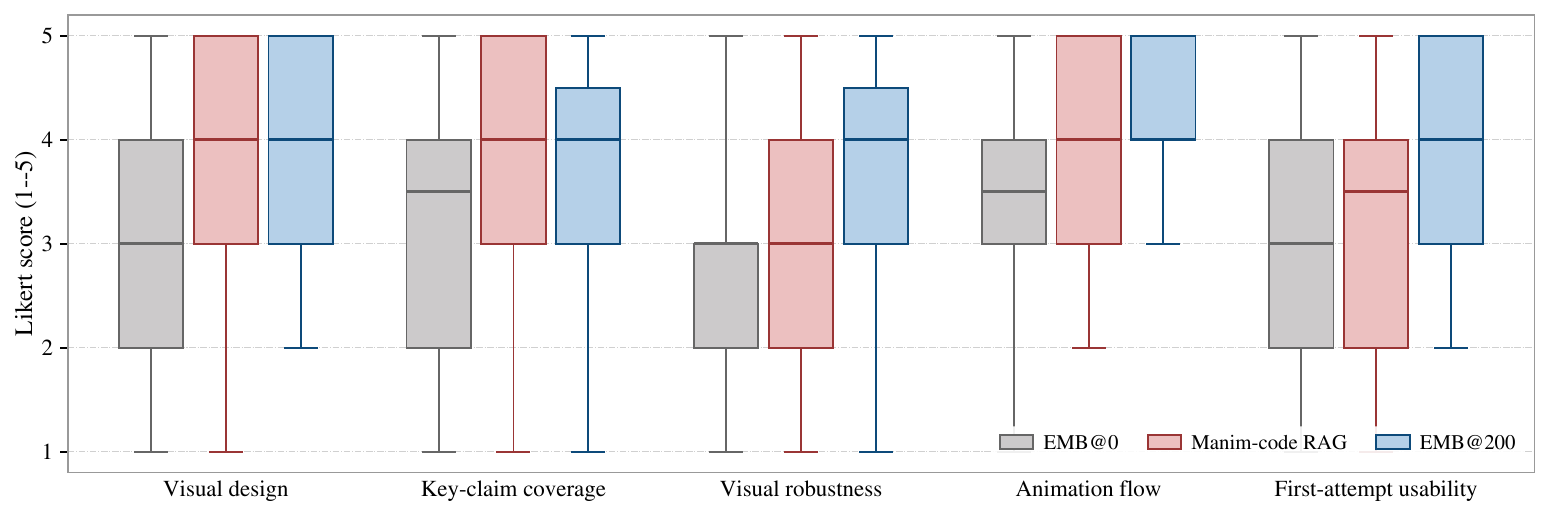}
\caption{\textbf{Per-rater Likert scores spread across the five human-evaluation dimensions for three representative conditions.}
Box distributions are derived from per-condition per-dimension rater scores
(EMB@0 $n{=}50$, Manim-code RAG $n{=}54$, EMB@200 $n{=}53$). EMB@200
dominates EMB@0 on every dimension and lifts the lower-quartile floor relative
to Manim-code RAG, indicating that the headline mean is not driven by a few
high-scoring~raters.}
\label{fig:likert-box}
\end{figure*}

\noindent Table~\ref{tab:human-scoring-fields} summarises aggregation of these
ratings. Human quality is the mean of the five dimensions, equally weighted and
averaged across raters. The dimensions are correlated but capture distinct
aspects: visual design and visual robustness concern visual presentation;
key-claim coverage concerns content coverage; animation flow concerns temporal
presentation; and first-attempt usability is a holistic readiness judgement.
Figure~\ref{fig:likert-box} shows the per-dimension Likert distribution for the
three anchor conditions (EMB@0, Manim-code RAG, EMB@200); the dimension-level
story matches the headline mean, with EMB@200 lifting both the median and the
lower-quartile floor over the two~baselines.

\paragraph{Human Pass@1.}
Raters answer a binary usability question: ``Is this first-attempt
video usable as a paper-section animation without major repair?'' A major repair
is defined as any change requiring structural reorganization of the scene plan or
a complete rewrite of the animation code. Human Pass@1 is computed from all
collected binary judgements within each condition. All submitted blind ratings
are retained without majority voting or post hoc adjudication.

\paragraph{Fatal flags.}
Raters additionally mark zero or more fatal issue flags from a closed set:
unrelated to the target, major factual or formula error, hallucinated claim,
missing central claim, severe overlap or occlusion, cropped or off-screen
object, unreadable text or labels, confusing animation order, overly static
presentation, excessive text, broken or unplayable output, and other. An
optional free-text note records additional observations; selecting other
requires such a description.

\begin{table}[t]
\centering
\small
\setlength{\tabcolsep}{4pt}
\begin{tabular}{p{0.58\linewidth}p{0.30\linewidth}}
\toprule
Human-scoring item & Aggregation \\
\midrule
Usability decision & proportion of individual pass votes \\
Visual design & mean score \\
Key-claim coverage & mean score \\
Visual robustness & mean score \\
Animation flow & mean score \\
First-attempt usability & mean score \\
Human quality & mean of five dimensions \\
Fatal issue flags & frequency \\
\bottomrule
\end{tabular}
\caption{\textbf{Human scoring combines a binary item, five Likert dimensions, and fatal flags.} Each rating includes a
binary usability judgement, dimensional ratings, and optional fatal-issue flags.}
\label{tab:human-scoring-fields}
\end{table}

\paragraph{Inter-rater agreement.}
Fleiss' $\kappa$ for binary usability votes is $-0.01$, and binary full
agreement is $59.4\%$. Intraclass correlation (ICC) for the five Likert
dimensions is low across all dimensions:
visual\_design ${=}\,0.24$, key\_claim\_coverage ${=}\,0.00$,
visual\_robustness ${=}\,0.10$, animation\_flow ${=}\,0.33$, and
first\_attempt\_usability ${=}\,{-0.09}$. All dimensions fall below common reliability
thresholds ($\kappa{\ge}0.4$, ICC${\ge}0.5$), reflecting the subjective nature
of animation quality assessment and the disciplinary diversity of the rater pool.
We retain all independent blind ratings without exclusion, majority voting, or
post hoc adjudication. Because the original rater--video-level records were
deleted after the 14-day retention period, condition-specific denominators and
paired uncertainty intervals cannot be reliably reconstructed from the retained
aggregate~results.

\section{VLM Scoring Schema and VLM--Human Agreement}
\label{sec:appendix-vlm-human}

The VLM scores rendered keyframes to drive revision and to gate memory writes.
Because the same VLM serves both roles, VLM scores are not used as headline
quality evidence. They are reported here as auxiliary diagnostics.

\paragraph{VLM scoring schema.}
Table~\ref{tab:vlm-dimensions} defines the three axes used by the VLM reviewer:
logical flow, layout/occlusion, and accuracy. The aggregate VLM score $u(t)$ is the unweighted
mean of the three per-axis scores, each on $[0,100]$. These scores drive visual
revision (the loop terminates when $u(t)\geq\theta_{\mathrm{conv}}=90$ or
$t=T_{\mathrm{vis}}=2$) and memory consolidation. The $\mathcal{M}^{+}$ write
threshold is set to $\theta_{\mathrm{high}}=85$ (slightly below the convergence
threshold, so that a scene can be judged good enough to remember without having
reached the auto-pass bar). The $\mathcal{M}^{-}$ write gate requires an
improvement of at least $\Delta_{\mathrm{fail}}=5$ between adjacent attempts. The
quality conclusions of the main paper are based on human ratings, not VLM scores.

\begin{table}[t]
  \centering
  \small
  \begin{tabular}{p{0.22\linewidth}p{0.66\linewidth}}
    \toprule
    \textbf{Dimension} & \textbf{What the VLM evaluates} \\
    \midrule
    Logical Flow       & Whether scene ordering follows the pedagogical narrative and animation
                         transitions are not abrupt. \\
    Layout / Occlusion & Whether elements overlap, exceed the canvas, or are hidden behind
                         formulae. \\
    Accuracy           & Whether formulae are correct and the visualisation does not invite
                         misinterpretation. \\
    Aggregate score    & Unweighted mean of the three axes; used only as an internal
                         repair and memory-write signal. \\
    \bottomrule
  \end{tabular}
  \caption{\textbf{Three VLM axes drive revision and memory writes.}
  The table defines the logical-flow, layout/occlusion, and accuracy dimensions
  scored on $[0,100]$. Their unweighted mean is a fixed internal control signal,
  not a substitute for the reported blind human quality metric.}
  \label{tab:vlm-dimensions}
\end{table}

\paragraph{Evaluator leakage and self-confirmation risk.}
The VLM that scores keyframes also gates memory writes, so systematic VLM biases
could be consolidated into the EMB as self-confirming patterns. We mitigate this
in three ways. (1)~Headline quality evidence is always blind human judgement,
stored in a separate file never shown to any system. (2)~The negative-channel
  write gate requires a locally improving adjacent attempt pair with a margin
  $\Delta_{\mathrm{fail}}=5$, providing an operational improvement filter.
  Here, ``validation'' means only that a revision receives a better internal
  VLM evaluation than its preceding version; it does not imply human validation
  or rigorous causal identification. (3)~The
weak-VLM ablation (Table~\ref{tab:ablations}) quantifies sensitivity to
the choice of reward model.

\paragraph{VLM axis independence.}
The three VLM axes are aggregated by unweighted mean, chosen a priori. We verify
the aggregate signal against blind human ratings after collection
(Table~\ref{tab:human-vlm}). Because aggregate agreement is weak, we do
not use this validation to claim either axis reliability or score calibration;
the axes remain internal heuristics designed to cover complementary animation
failure~modes.

\paragraph{Prompt sensitivity.}
Retrieval-query wording and prompt-slot formatting are fixed engineering
choices that we do not ablate. The weak-VLM condition in
Table~\ref{tab:ablations} bounds sensitivity to the reward model but cannot
eliminate reviewer blind spots shared across VLM variants.

\paragraph{Agreement analysis.}
The agreement analysis uses 37 videos for which both VLM scores and completed
three-rater human annotations are available after filtering missing or failed
renders. The VLM decision for each video is paired with its three human votes. We
compute Pearson $r$ and Spearman $\rho$ between aggregate VLM score and Human
quality, and Cohen's $\kappa$ between VLM pass/fail (thresholded at 90) and
individual human binary usability votes. The weak, slightly negative agreement values reported in
Table~\ref{tab:human-vlm} support using blind human ratings for headline claims
and treating VLM scores only as an internal repair/write~signal.

\begin{table}[!t]
\centering
\small
\setlength{\tabcolsep}{4pt}
\begin{tabular}{p{0.68\linewidth}p{0.18\linewidth}}
\toprule
Agreement statistic & Value \\
\midrule
Pearson $r$ (VLM aggregate vs.\ Human quality) & $-0.17$ \\
Spearman $\rho$ (VLM aggregate vs.\ Human quality) & $-0.20$ \\
Agreement rate (VLM pass/fail vs.\ Human Pass@1) & 0.49 \\
Cohen's $\kappa$ (VLM pass/fail vs.\ Human Pass@1) & $-0.07$ \\
\bottomrule
\end{tabular}
\caption{\textbf{VLM scores are not reliable headline evaluation evidence.}
On 37 paired videos, agreement with blind human ratings is weak or slightly
negative; VLM scores are therefore used only as an internal repair/write~
signal.}
\label{tab:human-vlm}
\end{table}

\section{EMB Schema}
\label{sec:appendix-emb-examples}

Table~\ref{tab:emb-schema} reports the field-level schema of the dual-channel
EMB referenced in \S\ref{sec:method-emb}: which fields are shared across
polarities, which differ, and how each channel is written and injected.

\begin{table}[t]
  \centering
  \small
  \setlength{\tabcolsep}{3pt}
  \begin{tabular}{p{0.20\linewidth}p{0.34\linewidth}p{0.34\linewidth}}
    \toprule
    \textbf{Component}    & \textbf{$\mathcal{M}^{+}$ (success)} & \textbf{$\mathcal{M}^{-}$ (failure)} \\
    \midrule
    context (shared)      & \multicolumn{2}{p{0.70\linewidth}}{$s$, $r$, $d$, paper id, section id} \\
    provenance (shared)   & \multicolumn{2}{p{0.70\linewidth}}{run id, scene id, source, ordinal, before/after scores} \\
    body                  & rationale, final code, frame hash    & trigger, root cause, fix recipe, anti/good example, diagnostic \\
    inject\,as            & Reference Examples (soft)            & Known Pitfalls (hard) \\
    write condition       & best candidate meets positive-memory score threshold (85) & render succeeds after error, or visual score improves by at least 5 points \\
    distilled by          & Rationale Writer ($T=0.3$)           & Lesson Distiller ($T=0$) \\
    \bottomrule
  \end{tabular}
  \caption{\textbf{The EMB separates shared context from channel-specific bodies and writes.}
  Indexing stays channel-agnostic (one SQLite table, one retrieval interface), while
  two per-polarity Faiss indices preserve the soft-template / hard-exclusion split. Lessons
  and rationales are produced by dedicated LLM distillers rather than stored as literal
  before/after diffs.}
  \label{tab:emb-schema}
\end{table}

\section{Coder Prompt Template}
\label{sec:appendix-coder-prompt}

The Coder prompt is assembled per scene by concatenating four blocks: fixed
system instructions, the top-$k_{+}$ retrieved positive memories formatted as
Reference Examples, the top-$k_{-}$ retrieved negative memories
formatted as Known Pitfalls, and the scene plan from the \textsc{Storyboarder}.
Retrieval depths ($k_{+}=2$, $k_{-}=3$) and length caps match
Table~\ref{tab:fixed-hparams}, and per-channel content is drawn from the
fields of Table~\ref{tab:emb-schema}. When either channel returns fewer than
the requested $k$ entries, the corresponding block is rendered empty, which
recovers the EMB@0 prompt structure exactly.

The Reference-Examples block is a soft template: the Coder may deviate when the
retrieved code does not fit the current scene. The Known-Pitfalls block is
treated as a hard exclusion list to be avoided verbatim.

Coder LLM: \texttt{GPT-5.5} (see Table~\ref{tab:fixed-hparams}).

\subsection*{System Prompt}

You are an expert Manim Community Edition v0.20 engineer. You write
\textbf{complete, runnable} Python files that render with
\texttt{manim render -ql} on a headless Linux box.

\paragraph{Output requirement (strict).}
Return \textbf{exactly one} Python code block fenced with \texttt{```python}
\ldots \texttt{```}. The block must define \textbf{one} class that subclasses
\texttt{Scene} whose name \textbf{exactly matches} the \texttt{name} field of
the requested scene; do not include \texttt{if \_\_name\_\_ ==
"\_\_main\_\_"}. No extra prose, no explanations, no second code block.

\paragraph{Mandatory imports / setup.}
Use \texttt{from manim import *}. Do not import \texttt{manim.opengl} or use
\texttt{--renderer=opengl} features. The renderer is~\texttt{cairo}.

\paragraph{Hard constraints.} Violations prevent~\mbox{rendering.}
\begin{itemize}
  \item The target host has \textbf{no display server}. Avoid OpenGL-only
  Mobjects (\texttt{OpenGLMobject}, \texttt{ThreeDScene} GPU effects).
  \item Use simple LaTeX. Allowed packages: \texttt{amsmath},
  \texttt{amssymb}, \texttt{mathtools}. Avoid exotic macros.
  \item No external image files. No fonts beyond what Manim ships with. No
  network calls.
  \item Keep total scene duration close to \texttt{duration\_hint}
  ($\pm 20\%$), and cap mobject count: prefer $\le 30$ simultaneously visible
  mobjects.
  \item Use \texttt{self.play(...)} for transitions and \texttt{self.wait(t)}
  between transitions.
\end{itemize}

\paragraph{Recommended patterns.}
\begin{itemize}
  \item Use \texttt{MathTex(r"a\^{}2 + b\^{}2 = c\^{}2")} (raw string) for
  equations and \texttt{Tex(r"...")} for prose with LaTeX.
  \item For colored emphasis, pass token lists into \texttt{MathTex} with the
  \texttt{color} keyword, or use \texttt{.set\_color\_by\_tex(...)}.
  \item Group with \texttt{VGroup(...)} and arrange via
  \texttt{.arrange(DOWN, buff=0.5)}; position via \texttt{.to\_edge(UP)},
  \texttt{.shift(...)}, or \texttt{.next\_to(...)}.
  \item Default animations: \texttt{Write}, \texttt{Create}, \texttt{FadeIn},
  \texttt{FadeOut}, \texttt{Transform}, \texttt{ReplacementTransform},
  \texttt{Indicate}. Default to plain \texttt{Scene}.
\end{itemize}

\paragraph{Self-check before responding.}
Class name matches the requested \texttt{scene.name}; every LaTeX backslash is
escaped (raw strings); a final \texttt{self.wait(...)} is present; no
\texttt{OpenGL*} or image / sound / file IO calls.

If a previous attempt is shown with an error, \textbf{fix the specific error
indicated}; do not change unrelated parts of the code.

\paragraph{Reference Examples and Known Pitfalls.}
You may receive two extra sections in the user message. \textit{Reference
  Examples} are past scenes that scored highly on the visual rubric; treat them
  as guidance, not as boilerplate. \textit{Known Pitfalls} are write-gated
failure$\to$success transitions from previous runs; each lists a trigger
pattern, root cause, an anti-example, and a good example, and the
anti-example pattern \textbf{must be avoided} in the output. Both sections
are optional and may be empty.

\subsection*{User Prompt Template}

\begin{quote}\ttfamily\small\raggedright
Scene to render:\\
\hspace*{1em}\{SCENE\_PLAN\}\\[2pt]
Reference Examples (may be empty):\\
\hspace*{1em}\{REFERENCE\_EXAMPLES\}\\[2pt]
Known Pitfalls (may be empty):\\
\hspace*{1em}\{KNOWN\_PITFALLS\}
\end{quote}

\noindent The placeholders are filled per scene: \texttt{\{SCENE\_PLAN\}} is
the storyboarder output (\texttt{name}, \texttt{claim}, \texttt{evidence},
\texttt{takeaway}, \texttt{duration\_hint}); \texttt{\{REFERENCE\_EXAMPLES\}}
holds $k_{+}{=}2$ entries from $\mathcal{M}^{+}$ (rationale plus code
excerpt); \texttt{\{KNOWN\_PITFALLS\}} holds $k_{-}{=}3$ entries from
$\mathcal{M}^{-}$ (trigger, root cause, fix recipe, anti-/good example pair).
Field schemas follow Table~\ref{tab:emb-schema}.

\section{Supplementary Ablations}
\label{sec:appendix-ablations}
\label{sec:exp-ablations}

This appendix provides the per-variant mechanism rationale and additional
interpretation for the ablation results reported in \S\ref{sec:rq2}. All
variants are evaluated on the same fixed probe in read-only~mode.

\paragraph{Channel isolation.}
Removing $\mathcal{M}^{+}$ or $\mathcal{M}^{-}$ individually tests whether the
two channels are complementary: if so, single-channel variants should fall
between the full EMB and the no-memory baseline rather than match either
extreme. Removing both recovers the no-memory baseline
(Table~\ref{tab:ablations}).

\paragraph{Write gates.}
Positive memories require a score of at least $\theta_{\mathrm{high}}=85$;
failure memories require either a repaired render crash or a visual-score
increase of at least $\Delta_{\mathrm{fail}}=5$. The ungated failure-memory
variant tests whether the gate is load-bearing: if storing every failed-attempt
  diagnostic preserves quality, the gate is decorative; if Pass@1 degrades, the
  improvement-margin requirement is doing real work in keeping $\mathcal{M}^{-}$
free of random VLM noise.

\paragraph{Retrieval depth.}
Top-$1$ retrieval tests whether one nearest neighbour per channel suffices, so
gains that survive top-$1$ retrieval cannot be attributed to context-padding
effects. Random EMB (Table~\ref{tab:main-results}) preserves record count,
format, and token budget while breaking content--query alignment, isolating
content from format.

\paragraph{Rationale-only success memory.}
This variant removes final code excerpts from success-memory retrieval while
retaining the LLM-distilled rationale, testing whether the positive-channel
benefit reduces to copying past code rather than to the higher-level lesson
encoded in the rationale.

\paragraph{Matched-token RAG baselines.}
We compare retrieval from Manim documentation alone with retrieval from both
documentation and code under the same budget, encoder, and index used for EMB.
The sweep characterises what kind of retrievable content suffices and
isolates the EMB gain from any static-corpus retrieval gain.

\paragraph{Weak-VLM sensitivity.}
Replacing the reviewer with a weaker VLM (Step-3.7-Flash) tests whether the
improvement depends on one particular internal scoring model: results that
survive a weaker reviewer cannot be attributed to the idiosyncrasies of a
single VLM. Blind human ratings remain the reported quality evidence.

\begin{figure*}[!t]
\centering
\includegraphics[width=\linewidth]{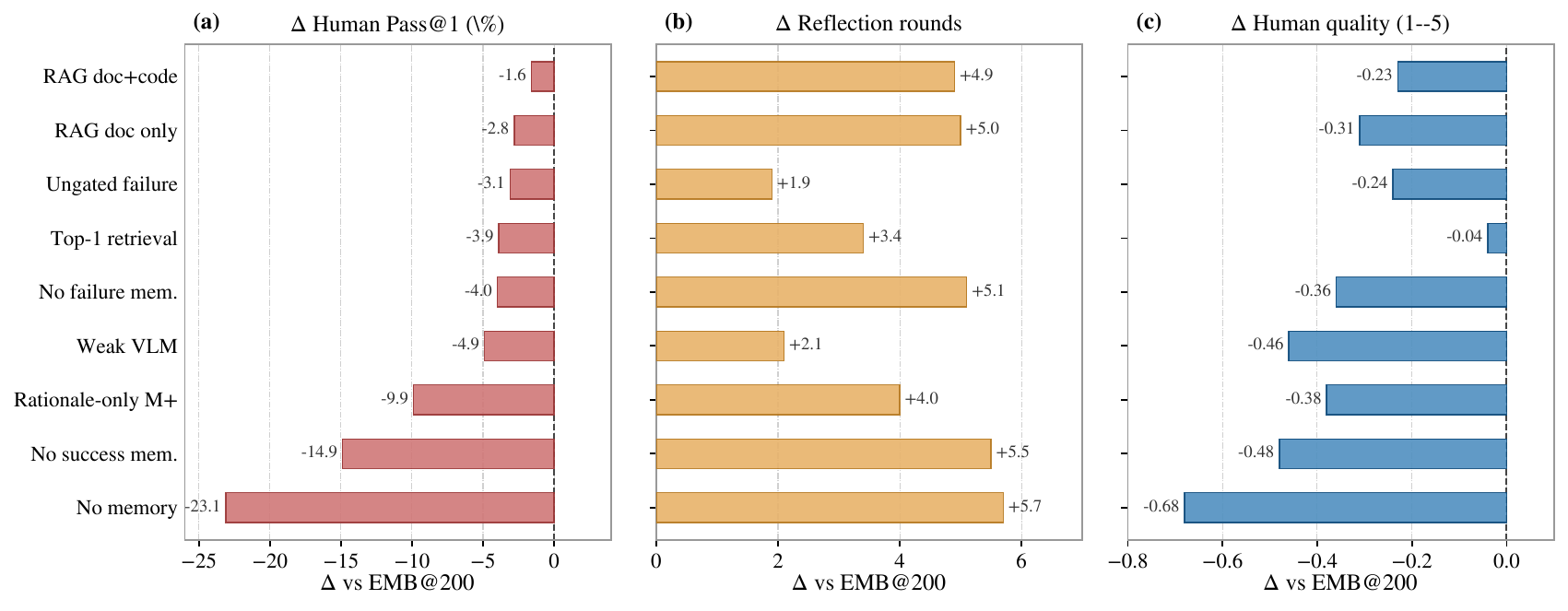}
\caption{\textbf{Per-ablation deltas measure the gap from each variant to Full EMB.} Each row is one variant of
Table~\ref{tab:ablations}; bars show the gap to Full EMB on each metric
(negative = worse for Pass@1 / quality, positive = worse for reflection
rounds, since fewer rounds is better). Removing $\mathcal{M}^{+}$ collapses
Pass@1 (left panel) more than removing $\mathcal{M}^{-}$, while removing
$\mathcal{M}^{-}$ inflates reflection rounds nearly as much as removing memory
entirely (middle panel), supporting the asymmetric channel-role reading in
\S\ref{sec:rq2}. Matched-budget RAG variants (top two rows) close most of the
Pass@1 gap but stay high in reflection rounds. Point deltas computed from Table~\ref{tab:ablations}.}
\label{fig:ablation-forest}
\end{figure*}

\paragraph{Forest-plot summary.}
Figure~\ref{fig:ablation-forest} aggregates the per-variant deltas in a single
panel: the channel asymmetry (removing $\mathcal{M}^{+}$ hurts Pass@1, removing
$\mathcal{M}^{-}$ inflates reflection rounds) is the most visible signal, and the
two matched-budget RAG variants close most of the Pass@1 gap but stay high in
reflection rounds.

\section{Snapshot Position Curve}
\label{sec:appendix-online-curve}

The fixed-probe snapshot experiment (\S\ref{sec:rq1}) is the headline result
because it controls for task-order effects. For completeness, we align its
four snapshot measurements with their positions in the memory-building stream.

\begin{figure}[!t]
\centering
\includegraphics[width=\compactfigwidth]{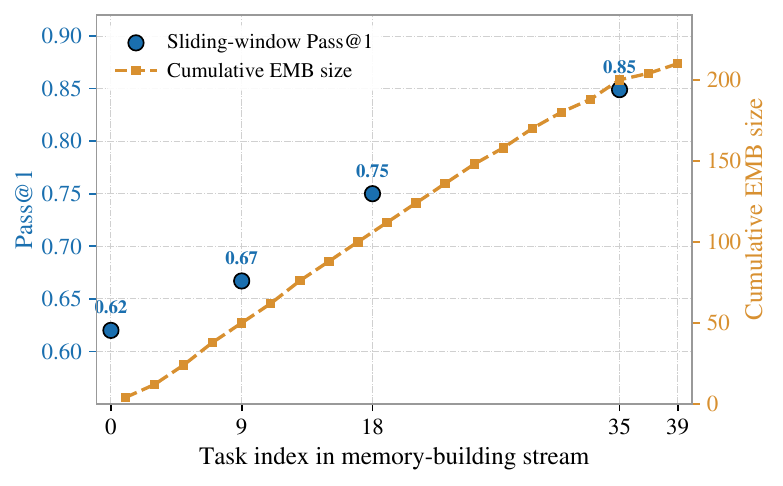}
\caption{\textbf{\textsc{ManimAgent} evolves online across the memory-building stream.} Left axis: Human Pass@1 at the four frozen snapshot points
(EMB@0, EMB@50, EMB@100, EMB@200). Right axis: cumulative EMB size
interpolated across the 39-task memory-building stream, anchored at the
four snapshot sizes and the final EMB size (${\approx}210$ records).}
\label{fig:online-curve}
\end{figure}

\paragraph{Protocol.}
\textsc{ManimAgent} processes the memory-building split sequentially. Before
each task, the current EMB is available for retrieval; after convergence, new
entries may be written back. The fixed-probe evaluation uses a separate run with
a frozen EMB snapshot. Figure~\ref{fig:online-curve} aligns fixed-probe Human Pass@1
at the four frozen snapshot sizes with the positions at which those snapshots
were taken in the memory-building stream.

\paragraph{Interpretation.}
The curve visualises when each frozen bank size was reached; its quality values
remain fixed-probe measurements rather than online evaluations of
memory-building tasks. The fixed-probe snapshot result is the primary evidence.

\section{Qualitative Inspection of Retrieved Memories}
\label{sec:appendix-qualitative}

An exploratory probe of what kind of information the EMB stores. This is not
headline evidence.

We manually label the top-$k$ retrieved $\mathcal{M}^{+}$ entries for a
stratified sample of probe queries, categorizing each as domain-specific,
layout/pacing/formula-rendering, or mixed. In this inspection, $83.3\%$ of
top-$k$ matches concern layout, pacing, or formula rendering, $16.7\%$ are
mixed, and domain-specific matches are absent, indicating that the positive
channel primarily transfers visual-presentation lessons rather than
domain-specific content.

\section{Qualitative Trace Analysis}
\label{sec:appendix-qualitative-trace}

We additionally inspected one complete test run directory for the Introduction
section of a BERT paper source (\texttt{arxiv-src:1810.04805}). This inspection
is a trace-level diagnostic, not headline evidence: it uses the saved
\texttt{summary.json}, \texttt{storyboard.json}, \texttt{trace.jsonl}, rendered
keyframe montages, and per-scene render manifests to understand the strengths
of the generated paper-section animation.

The summarizer extracted the concepts emphasized by the source introduction:
bidirectional pre-training, masked language modeling, next-sentence prediction,
the Transformer encoder, and fine-tuning. The storyboard converted these
concepts into a teaching sequence that introduces BERT, contrasts
left-to-right and bidirectional context, visualizes masked-token prediction,
and closes with a fine-tuning takeaway. The final rendered output gives a
compact visual explanation of how bidirectional context and reusable
representations support downstream NLP tasks.

Table~\ref{tab:bert-trace} summarises the four rendered scenes, pairing the
first-attempt and final VLM axis scores with a short qualitative strength of
each delivered~render.

\begin{table*}[t]
\centering
\footnotesize
\setlength{\tabcolsep}{4pt}
\begin{tabular}{@{}p{0.16\linewidth}p{0.25\linewidth}p{0.15\linewidth}p{0.36\linewidth}@{}}
\toprule
Scene & Trace outcome & VLM axes & Qualitative strength \\
\midrule
Title & Rendered; two visual revisions; final pass &
$78/58/94 \rightarrow 92/91/95$ &
The final version splits the long title across two balanced lines, centers the
encoder stack, and presents the two core cues of BERT---bidirectional context and
fine-tuning---as an immediately readable overview. \\
Bidirectional context & Rendered and passed visual review & $92/88/96$ &
The left-to-right arrows and two-sided mask context directly match the
bidirectionality claim of the paper, making the contrast between unidirectional and
bidirectional pre-training visually explicit. \\
Masked LM & Rendered; two visual revisions & $72/83/58 \rightarrow 84/88/62$ &
The revised scene groups surrounding context with clear green arrows and
highlights the predicted token, turning the masked-language-modeling objective
into a step-by-step visual explanation. \\
Fine-tuning takeaway & Rendered; two visual revisions &
$82/76/88 \rightarrow 88/82/86$ &
Larger task boxes, a task-count cue, and a boxed takeaway improve the hierarchy
from pre-trained BERT to downstream heads and make the final message easier to
read. \\
\bottomrule
\end{tabular}
\caption{\textbf{One BERT test run shows positive qualitative progress across scenes.}
Scores are reported as logical flow / layout-occlusion / accuracy for the first
and final renderable versions recorded in the trace.}
\label{tab:bert-trace}
\end{table*}

\begin{figure*}[t]
\centering
\rmfamily\footnotesize
\setlength{\tabcolsep}{4pt}
\renewcommand{\arraystretch}{1.05}
\begin{tabular}{@{}>{\centering\arraybackslash}m{0.13\linewidth}|>{\centering\arraybackslash}m{0.83\linewidth}@{}}
\textbf{Scene / version} & \textbf{Rendered keyframe montage} \\
\midrule
\shortstack[c]{Title\\first} &
\includegraphics[width=\linewidth,height=0.11\textheight,keepaspectratio]{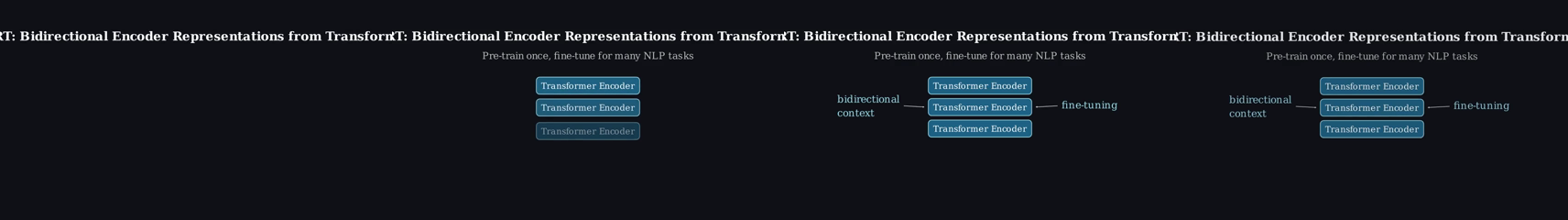} \\
\cmidrule(lr){2-2}
\shortstack[c]{Title\\final} &
\includegraphics[width=\linewidth,height=0.11\textheight,keepaspectratio]{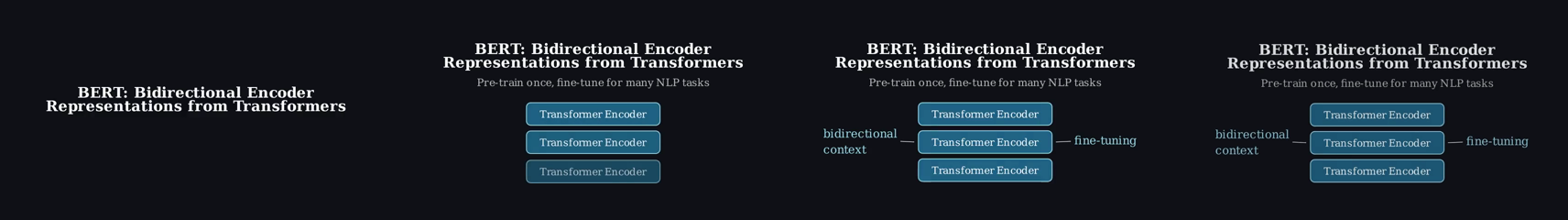} \\
\midrule
\shortstack[c]{Masked LM\\first} &
\includegraphics[width=\linewidth,height=0.11\textheight,keepaspectratio]{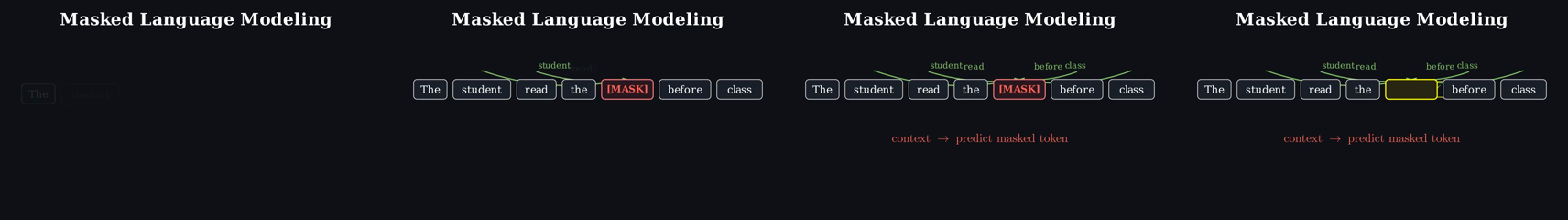} \\
\cmidrule(lr){2-2}
\shortstack[c]{Masked LM\\final} &
\includegraphics[width=\linewidth,height=0.11\textheight,keepaspectratio]{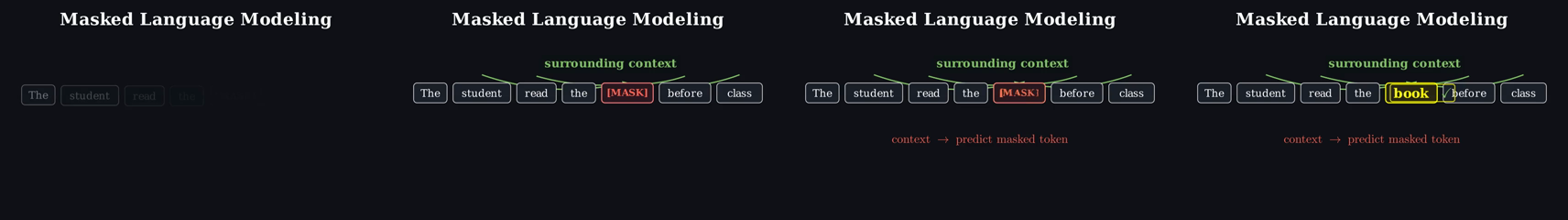} \\
\midrule
\shortstack[c]{Fine-tuning\\first} &
\includegraphics[width=\linewidth,height=0.11\textheight,keepaspectratio]{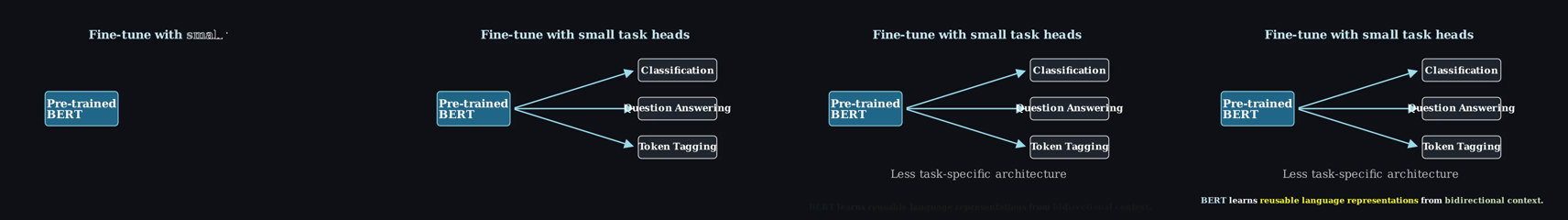} \\
\cmidrule(lr){2-2}
\shortstack[c]{Fine-tuning\\final} &
\includegraphics[width=\linewidth,height=0.11\textheight,keepaspectratio]{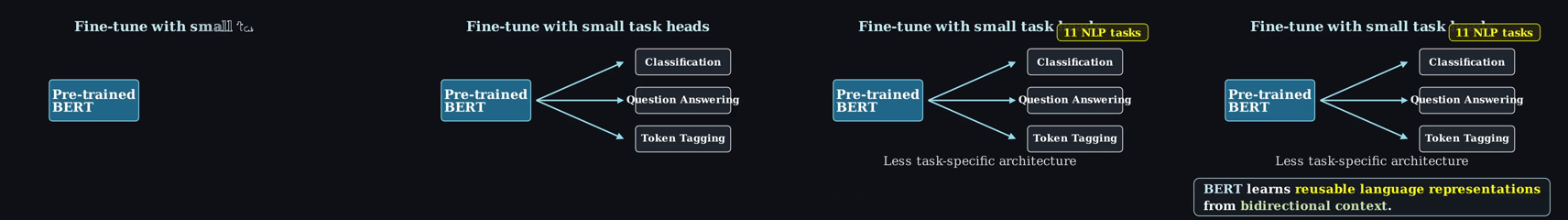} \\
\end{tabular}
\caption{\textbf{Detailed qualitative visual revisions in one BERT test run improve readability across scenes.}
This appendix figure keeps the original long-strip keyframe montages at a
readable width. It compares the first renderable version and final recorded
version for three rendered scenes: the title scene becomes a balanced overview,
the masked-language-modeling scene makes context aggregation and token
prediction clearer, and the fine-tuning scene gives the downstream-task
hierarchy and final takeaway stronger visual emphasis.}
\label{fig:bert-qualitative-revisions}
\end{figure*}

This trace shows how the pipeline turns an abstract NLP introduction into a
coherent visual narrative. The title scene establishes the model identity and
main conceptual hooks; the bidirectional-context scene makes the central
architectural contrast concrete through opposing arrow patterns; the
masked-language-modeling scene animates the pre-training objective as a visible
prediction process; and the fine-tuning scene connects the shared encoder to
multiple downstream task heads. Across the rendered scenes (Figure~\ref{fig:bert-qualitative-revisions}), visual reflection
mainly improves readability: long text is broken into stable blocks, arrows and
labels become easier to parse, and the final takeaway is given stronger visual
emphasis.

\section{Reproducibility Details}
\label{sec:appendix-reproducibility}

Table~\ref{tab:fixed-hparams} lists the fixed hyperparameters used in the main
experiment. The two VLM thresholds ($\theta_{\mathrm{high}}=85$,
$\theta_{\mathrm{conv}}=90$) and the failure-channel improvement margin
($\Delta_{\mathrm{fail}}=5$) are the only memory-write knobs; the retrieval
depths ($k_{+}=2$, $k_{-}=3$) and snapshot sizes $\{0,50,100,200\}$ are pinned
across all EMB conditions.

Table~\ref{tab:cost} reports approximate per-task resource use. LLM and VLM
token budgets are stable across conditions (within roughly $\pm$10\%); the
main cost gradient is wall-clock render time, which rises from 19 minutes at
EMB@0 to 31 minutes at EMB@200 as retrieved scenes prompt more elaborate
renders.

\paragraph{Deployment-scale memory management.}
We evaluate the EMB only through 400 entries ($\mathcal{O}(10^{2})$).
Production-scale retrieval ($\mathcal{O}(10^{6})$) is likely to require
hierarchical indexing, semantic deduplication, consolidation of similar
memories, utility-based eviction, and dynamic forgetting. Write gates defined
on the current VLM scores also do not prevent memory rot after an LLM or VLM
upgrade, and per-task cost continues to depend on the reflection budgets.

A complete run manifest records: the LLM and VLM
model identifiers with version tags, max output tokens, retrieval configuration ($k_{+}$, $k_{-}$, index type, embedding
model), renderer version, random seed, evaluation split, and ablation preset.

\begin{table}[!h]
  \centering
  \small
  \setlength{\tabcolsep}{4pt}
  \begin{tabular}{lccc}
    \toprule
    \textbf{Condition} & \shortstack{LLM\\tokens} & \shortstack{VLM\\tokens} & \shortstack{Render\\time} \\
    \midrule
    Manim-code RAG       & 38K & 23K & 27 min \\
    Random EMB           & 38K & 23K & 25 min \\
    EMB@0                & 37K & 23K & 19 min \\
    EMB@100              & 37K & 22K & 23 min \\
    EMB@200              & 36K & 20K & 31 min \\
    \bottomrule
  \end{tabular}
  \caption{\textbf{Per-task cost stays comparable across conditions.} Token counts estimated from output
  characters and review-call counts; no explicit token tracking was enabled. Render
  time is wall-clock duration per task.}
  \label{tab:cost}
\end{table}

\begin{table*}[!t]
\centering
\footnotesize
\setlength{\tabcolsep}{5pt}
\renewcommand{\arraystretch}{1.15}
\begin{minipage}[t]{0.49\linewidth}
\centering
\begin{tabular}{@{}lr@{}}
\toprule
Parameter & Value \\
\midrule
Convergence threshold (auto-pass) & 90 \\
$\mathcal{M}^{+}$ write threshold & 85 \\
$\mathcal{M}^{-}$ improvement margin & 5 \\
Text-reflection retry budget $T_{\mathrm{text}}$ & 2 \\
Visual-reflection revision budget $T_{\mathrm{vis}}$ & 2 \\
Positive retrieval depth $k_{+}$ & 2 \\
Negative retrieval depth $k_{-}$ & 3 \\
$\mathcal{M}^{+}$ code-excerpt cap & 1200 chars \\
\bottomrule
\end{tabular}
\end{minipage}\hfill
\begin{minipage}[t]{0.49\linewidth}
\centering
\begin{tabular}{@{}lr@{}}
\toprule
Parameter & Value \\
\midrule
$\mathcal{M}^{+}$ rationale cap & 600 chars \\
Snapshot sizes $\mathcal{K}$ & $\{0,50,100,200\}$ \\
Sentence encoder $\phi$ & all-MiniLM-L6-v2 \\
Vector index & Faiss IndexFlatIP \\
Coder LLM & GPT-5.5 \\
Reviewer VLM & GPT-5.5 \\
Weak-VLM ablation & Step-3.7-Flash \\
\textsc{Manim} version & Community v0.20.1 \\
\bottomrule
\end{tabular}
\end{minipage}
\caption{\textbf{All numerical hyperparameters are pinned across conditions.} Encoder $\phi$ outputs 384-dimensional
vectors; the Faiss index applies inner product to $L_{2}$-normalised
embeddings (equivalent to cosine similarity). Caps are enforced at
retrieval time. Values match code defaults and run manifests.}
\label{tab:fixed-hparams}
\end{table*}

\paragraph{Pre- vs.\ post-cutoff stratification.}
To assess whether gains are driven by memorised paper text rather than by EMB
retrieval, Table~\ref{tab:cutoff-stratification} stratifies probe tasks by
publication date relative to the training cutoff of the LLM. A large gap favouring pre-cutoff papers would suggest memorised
source text; a small or non-systematic gap supports EMB-mediated transfer.
The observed deltas (Pass@1 $-5.9\%$ pre minus post, quality $+0.17$) point in
opposite directions and are small, ruling out a consistent pre-cutoff advantage
and so supporting the EMB-mediated reading.

\par\smallskip\noindent\begin{minipage}{\columnwidth}
\centering
\small
\setlength{\tabcolsep}{4pt}
\begin{tabular}{lcc}
\toprule
Stratum & Human Pass@1 & Human quality \\
\midrule
Pre-cutoff papers & 74.1\% & 3.56 \\
Post-cutoff papers & 80.0\% & 3.40 \\
$\Delta$ (pre $-$ post) & -5.9\% & +0.17 \\
\bottomrule
\end{tabular}
\captionsetup{type=table,hypcap=false}
\caption{\textbf{Stratifying by LLM training cutoff shows no consistent pre-cutoff advantage.} A small gap supports
EMB-mediated transfer over memorization.}
\label{tab:cutoff-stratification}
\end{minipage}\par\smallskip

\paragraph{Manim-code RAG corpus.}
The fixed corpus contains 360 Manim API reference entries, 10 standalone
code examples from the official examples page, and one quick-start tutorial
excerpt, totalling approximately 370,000 characters (54,000 tokens). The API
documentation mainly covers \texttt{manim.mobject}, \texttt{manim.utils},
\texttt{manim.animation}, \texttt{manim.scene}, and \texttt{manim.camera};
the examples include coordinate axes, bar charts, equation derivations, and
three-dimensional surfaces. Median entry length is 629 characters and the
maximum is 6,000. Each entry records its source provenance and MIT License,
with original notices and attribution preserved. Retrieval uses the same
encoder, Faiss vector-index configuration, and context budget as the EMB.

\section{Retrieval-to-Revision Case Study}
\label{sec:appendix-retrieval-case}

We trace one fixed-probe example from memory retrieval to code revision. The
task is to animate self-attention over image tokens: the scene contains a row
of token rectangles labelled $x_1$ through $x_6$, curved attention arrows, a
Transformer encoder stack, and a formula. In the initial render, the dense
token row and numerous curved arrows are difficult to read: token labels are
small, the arrows are thin and visually uniform, and the attention arcs do not
stand out from the rest of the scene.

\paragraph{Retrieved memories.}
The positive channel retrieves success record \texttt{14872d25} with cosine
similarity 0.781 from \texttt{arxiv:2010.11929 / Introduction /
TransformerAttention}; its internal VLM score is 87.0. The record describes a
successful left-to-right layout that separates tokens, attention arcs, the
encoder stack, and the formula. It also recommends staged
\texttt{FadeIn}/\texttt{Create} animation and high-contrast yellow--green arcs
above blue tokens to make global attention readable.

The negative channel retrieves failure record \texttt{2c4c3609} with cosine
similarity 0.781 from the same \texttt{TransformerAttention} task. This record
captures a revision whose internal VLM score increased from 75.33 to 87.0. Its
trigger is a dense token row that must be read together with many curved arrows
and the encoder stack or formula. The diagnosed failure pattern is small token
labels combined with uniform thin arcs; the stored remedy is to enlarge the
tokens and text, thicken the arrows, vary their curvature, and keep the arcs
above the background with a higher z-index.

\paragraph{Resulting code revision.}
After these memories are injected into the reflection prompt, the system makes
the following representative changes to the Manim code:

{\footnotesize
\begin{verbatim}
- rect = RoundedRectangle(width=0.72, height=0.58,
-                         stroke_width=3)
- tex = MathTex(label, font_size=28)
- arc = CurvedArrow(a, b, angle=-PI/2.35,
-                   stroke_width=3)
- arc.set_z_index(1)
+ rect = RoundedRectangle(width=0.92, height=0.70,
+                         stroke_width=3.2)
+ tex = MathTex(label, font_size=34)
+ for k, (i, j) in enumerate(conns):
+     arc = CurvedArrow(a, b,
+         angle=-PI/(2.25 + 0.12*k),
+         stroke_width=3.6)
+     arc.set_z_index(2)
\end{verbatim}
}

The retrieved $\mathcal{M}^{+}$ entry supplies the left-to-right organization,
staged animation, and high-contrast attention arcs, while the
$\mathcal{M}^{-}$ entry helps avoid small labels, uniform thin arcs, and weak
visual separation between attention links and the rest of the scene. After the
revision, the internal VLM score increases from 75.33 to 87.0, with clearer
layout, labels, and attention-link salience. This is an internal trace-level
diagnostic rather than an independently human-rated result.

This example also illustrates the memory representation: the EMB does not
store complete historical code or full conversations. It stores distilled
positive and negative experiences, retrieves those relevant to the current
task, and uses them to guide code reflection and revision.

\end{document}